\documentclass[journal,compsoc,cspaper,final]{IEEEtran}
\usepackage{url}
\usepackage{mathtools} 
\usepackage{amsmath,amsfonts,amssymb}
\usepackage{algorithmic}
\usepackage{graphicx}
\usepackage{mathrsfs}
\usepackage{amsbsy}
\usepackage{amsthm}
\usepackage{amssymb}
\usepackage{booktabs}
\usepackage[table]{xcolor}
\usepackage{textcomp}
\usepackage{bm}
\usepackage{ragged2e}

\usepackage{tikz}
\usepackage{tcolorbox}
\usetikzlibrary{arrows,shapes,chains}
\usetikzlibrary{spy}
\usepackage{diagbox} 
\usepackage{enumitem}
\usepackage{soul,xcolor} 
\usepackage[switch]{lineno} 
\usepackage{balance}
\usepackage{subfigure}
\usepackage{stfloats}
\usepackage{multirow}
\usepackage{xspace}
\usetikzlibrary{intersections}
\usepackage{caption}
\usepackage[nocompress]{cite}
\usepackage{algorithm}
\usepackage{makecell}
\usepackage{pifont}
\usepackage{bbding}
\usetikzlibrary{fit}
\usepackage{comment}
\usepackage{tabularx}
\usepackage[pagebackref,breaklinks,colorlinks]{hyperref}
\usepackage{tablefootnote}
\usepackage[export]{adjustbox}
\usepackage{boxedminipage}
\usepackage{colortbl}
\usepackage{pgfkeys}
\usepackage{flushend} 
\usepackage[capitalize]{cleveref}
\crefname{section}{Sec.}{Secs.}
\Crefname{section}{Section}{Sections}
\Crefname{table}{Table}{Tables}
\crefname{table}{Tab.}{Tabs.}
\usepackage{multirow}
\usepackage{overpic}
\newcommand{\ie}{\textit{i.e.}}
\newcommand{\eg}{\textit{e.g.}}
\newcommand{\etal}{\textit{et al.}}

\newcommand{\vs}{\textit{vs.}}

\newcommand{\myname}{EVDI++} %
\newcommand{\re}[1]{\textcolor{black}{#1}}

\begin{document}

\title{EVDI\text{++}: Event-based Video Deblurring and Interpolation via Self-Supervised Learning}

\author{Chi Zhang, Xiang Zhang, Chenxu Jiang, Gui-Song Xia, Lei Yu
\IEEEcompsocitemizethanks{
\IEEEcompsocthanksitem C. Zhang is with the School of Electronic Information, Wuhan University, Wuhan, 430072, China, and also with the Peng Cheng Laboratory, Shenzhen, 518000, China. E-mail: zhangchi1@whu.edu.cn.
\IEEEcompsocthanksitem X. Zhang is with the Computer Graphics Lab of ETH Zurich, Switzerland. E-mail: xiang.zhang@vision.ee.ethz.ch.
\IEEEcompsocthanksitem C. Jiang is with the School of Electronic Information, Wuhan University, Wuhan 430072, China. E-mail: jiangcx@whu.edu.cn.
\IEEEcompsocthanksitem G. Xia and L. Yu are with the School of Artificial Intelligence, Wuhan University, Wuhan, 430072, China. E-mail: \{guisong.xia, ly.wd\}@whu.edu.cn.
\IEEEcompsocthanksitem The research was partially supported by the National Natural Science Foundation of China under Grant 62271354 and the Fundamental Research Funds for the Central Universities.
\IEEEcompsocthanksitem C. Zhang and X. Zhang contributed equally to this work. 
\IEEEcompsocthanksitem Corresponding author: L. Yu.
}
}

\markboth{Submission to IEEE TPAMI}%
{Submission to IEEE TPAMI}


\IEEEtitleabstractindextext{
\justifying
\begin{abstract}
Frame-based cameras with extended exposure times often produce perceptible visual blurring and information loss between frames, significantly degrading video quality. To address this challenge, we introduce EVDI++, a unified self-supervised framework for Event-based Video Deblurring and Interpolation that leverages the high temporal resolution of event cameras to mitigate motion blur and enable intermediate frame prediction. Specifically, the Learnable Double Integral (LDI) network is designed to estimate the mapping relation between reference frames and sharp latent images. Then, we refine the coarse results and optimize overall training efficiency by introducing a learning-based division reconstruction module, enabling images to be converted with varying exposure intervals. We devise an adaptive parameter-free fusion strategy to obtain the final results, utilizing the confidence embedded in the LDI outputs of concurrent events. A self-supervised learning framework is proposed to enable network training with real-world blurry videos and events by exploring the mutual constraints among blurry frames, latent images, and event streams. We further construct a dataset with {\it real-world} blurry images and events using a DAVIS346c camera, demonstrating the generalizability of the proposed EVDI++ in real-world scenarios. Extensive experiments on both synthetic and real-world datasets show that our method achieves state-of-the-art performance in video deblurring and interpolation tasks. 
\par
{\noindent\bf Project page:} \url{https://bestrivenzc.github.io/EVDI-plus-plus/}.
\end{abstract}

}
\maketitle

\section{Introduction}
\IEEEPARstart{C}{hallenges} arise in high-quality video acquisition when dealing with highly dynamic scenes, \eg, fast-moving targets or complex non-linear motions, which stem from the common occurrence of frame blurring and the absence of essential target information between consecutive frames~\cite{telleen2007synthetic}. Current frame-based methods address these issues through motion deblurring~\cite{LEVS_jin2018learning,zamir2021multi}, frame interpolation~\cite{dain_bao2019depth,huang2022real}, and blurry video enhancement techniques~\cite{flawless_jin2019learning,bin_shen2020blurry}. However, frame-based deblurring methods face a formidable challenge in predicting sharp latent frames from severely blurred videos owing to the motion ambiguities and the erasure of intensity textures~\cite{LEVS_jin2018learning}. Moreover, the general assumption in existing frame-based interpolation methods~\cite{dain_bao2019depth} is that there is linear motion between adjacent frames, which often leads to inaccurate predictions in real-world scenarios, especially when dealing with complex non-linear motions.
\par

\begin{figure*}[!t]
  \centering
  \includegraphics[width=0.98\linewidth]{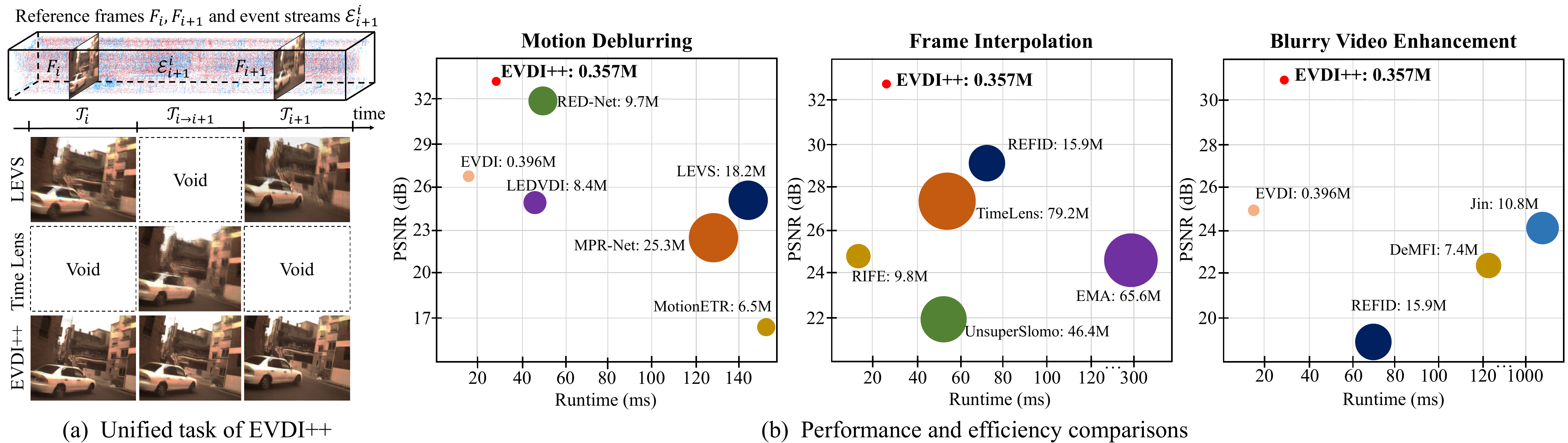}
    \vspace{-2mm}
  \caption{(a) Illustrative examples of video deblurring and interpolation using the state-of-the-art methods LEVS \cite{LEVS_jin2018learning} for deblurring and TimeLens \cite{timelens_tulyakov2021time} for interpolation, alongside our unified approach \myname, highlight the superior capabilities of \myname. Our method effectively reconstructs latent sharp images both within frames (intra-frame) and between frames (inter-frame) simultaneously, whereas the individual deblurring and interpolation methods are limited to estimating sharp images either intra-frame or inter-frame respectively. (b) Performance versus complexity diagram on the ColorDVS dataset for motion deblurring, frame interpolation, and blurry video enhancement sequence reconstruction tasks, respectively. The model in the top-left corner demonstrates the highest effectiveness, with each blob’s size proportional to the number of network parameters.}
  \label{Overview}
  \vspace{-5mm}
\end{figure*}

\re{Recent research has highlighted the advantages of event cameras~\cite{survey_9138762} in motion deblurring~\cite{edi_pan2019bringing,pan2020high,ledvdi_lin2020learning,esl_wang2020event,red_xu2021motion,zhang2024crosszoom,chen2024motion,xu2025motion,cho2023non,kim2023event,kim2024frequency} and frame interpolation~\cite{ledvdi_lin2020learning,timelens_tulyakov2021time,yang2024latency,cho2024tta}. On the one hand, event cameras inherently capture precise motion cues and sharp edges~\cite{benosman2013event} due to their asynchronous data acquisition with extremely low latency (on the order of $\mu s$)~\cite{lichtsteiner128Times1282008,survey_9138762}, which helps to effectively mitigate motion blur~\cite{edi_pan2019bringing,pan2020high,ledvdi_lin2020learning,esl_wang2020event,red_xu2021motion,chen2024motion,xu2025motion}. On the other hand, their ability to continuously capture brightness changes allows them to recover missing information between frames, enabling accurate intermediate frame reconstruction even in scenarios with complex, non-linear motion~\cite{edi_pan2019bringing,pan2020high,ledvdi_lin2020learning,esl_wang2020event,red_xu2021motion}. Despite these benefits, existing methods still face two major challenges in real-world scenarios:      
\begin{itemize}
\setlength{\itemindent}{0em}
    \item {\it Limitations of Separate Tasks:} Interpolation methods~\cite{timelens_tulyakov2021time,cho2024tta} heavily rely on the quality of reference frames, which poses difficulties when the inputs are severely degraded by motion blur. For deblurring, most approaches~\cite{esl_wang2020event,red_xu2021motion,chen2024motion,xu2025motion} focus solely on recovering sharp images within the exposure period of blurry frames, often ignoring latent images that lie temporally between blurry observations (see Fig.\ref{Overview} (a)).
    \item {\it Data Inconsistency:} Most prior works rely on labeled synthetic datasets for supervised\cite{esl_wang2020event,timelens_tulyakov2021time} or semi-supervised learning~\cite{red_xu2021motion}, since acquiring ground-truth labels for motion deblurring in highly dynamic real-world scenes is extremely difficult. However, models trained on synthetic data often suffer from performance degradation in real-world applications due to discrepancies between synthetic and real data distributions~\cite{red_xu2021motion}, as well as the scarcity of accurate ground-truth annotations.
\end{itemize}
}

\par

\re{These limitations make it evident that treating deblurring and interpolation as two independent tasks is suboptimal. While one might consider applying deblurring followed by interpolation (or vice versa) in a sequential manner, such cascaded strategies often lead to performance loss due to error accumulation across stages. To address this, a joint estimation strategy is needed that enables bidirectional information exchange between deblurring and interpolation to reduce cumulative errors. Furthermore, a self-supervised learning framework is crucial for alleviating domain gap issues and addressing the challenge of insufficient labeled data in real-world settings.} 

To this end, this paper proposes a unified self-supervised framework for Event-based Video Deblurring and Interpolation (\myname) to achieve high-quality video acquisition. The proposed method consists of three modules: a Learnable Double Integral (LDI) network, a Learning-based Division Reconstruction (LDR) module, and an Adaptive Parameter-free Fusion (APF) module. Specifically, the LDI network is designed to automatically predict the mapping relation between blurry frames and sharp latent images corresponding to the events, where the timestamp of the latent image can be chosen arbitrarily within the exposure time of the reference frames (deblurring task) or between consecutive reference frames (interpolation task). The LDR module is designed to refine the coarse results from the LDI network. Furthermore, the APF module receives the reconstruction of latent images and generates final results by utilizing the confidence embedded in the LDI outputs of concurrent events. For training, we leverage the mutual constraints among blurry frames, sharp latent images, and event streams, and propose a fully self-supervised learning framework to enable the network to fit the distribution of real-world data without requiring ground-truth images.


\par
The main contributions of this paper are three-fold:
\begin{itemize}
    \item We propose a unified framework for Event-based Video Deblurring and Interpolation, \ie, \myname, which can generate arbitrarily high frame-rate sharp videos from low frame-rate sharp or blurry inputs.
    \item We propose a fully self-supervised learning framework to directly fit real-world data distributions without requiring the collection of labeled datasets.
    \item Experiments on both synthetic and real-world datasets show that our \myname\ achieves state-of-the-art performance while maintaining an efficient network design and a compact model size.
\end{itemize}

This article is an extended version of our preliminary work, \ie, EVDI~\cite{zhang2022unifying}, with several significant improvements: (i) a Learning-based Division Reconstruction (LDR) module is further devised after the Learnable Double Integral (LDI) network to refine the coarse results by converting images with different exposure intervals; (ii) an Adaptive Parameter-free Fusion (APF) is proposed to efficiently select the most informative features from different inputs, resulting in more robust performance and fewer network parameters than its previous version, \ie, EVDI; (iii) we modify the self-supervised loss function composition to improve overall training speed and stability. Compared to EVDI, EVDI++ achieves an average PSNR improvement of 2.97 dB with reduced parameters (0.357M \vs\ 0.396M) and exhibits a 16-fold improvement in training speed. The performance and efficiency comparisons between our unified EVDI++ and state-of-the-art MD, FI, and BVE approaches are shown in \cref{Overview} (b).          

\section{Related Work}\label{Sec:RelatedWork}

\subsection{Motion Deblurring}\label{subSec:Deblurring}
One of the most popular frame-based deblurring methods is to employ neural networks to learn the blur feature and predict sharp images from blurry inputs \cite{LEVS_jin2018learning,8online_hyun2017online,42filter_zhou2019spatio,22intra_nah2019recurrent,zhong2022animation,zhong2023blur,zhong2023real}. Techniques such as generative adversarial training~\cite{kupyn2018deblurgan,kupyn2019deblurgan}, coarse-to-fine pyramid structure~\cite{tao2018scale,22intra_nah2019recurrent,zhang2019deep} are commonly applied to single-image deblurring tasks, substantially enhancing the reconstruction of latent sharp images. Additionally, various strategies have been devised to leverage temporal information within blurry frames, including a dynamic temporal blending mechanism \cite{8online_hyun2017online}, spatiotemporal filter adaptive networks \cite{42filter_zhou2019spatio}, and intra-frame iterations \cite{22intra_nah2019recurrent}. Despite these advances, the aforementioned methods primarily employ data-driven or optimization-based approaches on blurry inputs, without utilizing crucial intra-frame motion and texture, resulting in suboptimal performance and limited model generalization. 

Recent works have revealed the potential of events in motion deblurring. Event streams inherently embed motion information and sharp edges, which can be exploited to tackle the temporal ambiguity and texture erasure caused by motion blur. Pioneer event-based methods achieve motion deblurring by relating blurry frames, sharp latent images, and the corresponding events according to the physical event generation model \cite{edi_pan2019bringing,pan2020high}, but their performance is often degraded due to the imperfection of physical circuits, \eg, intrinsic camera noises. To alleviate this, learning-based approaches \cite{esl_wang2020event,red_xu2021motion,yu2023esl_plusplus} have been proposed to fit the distribution of event data, achieving better deblurring performance. To solve the aforementioned issues, learning-based approaches have emerged, significantly reducing the impact of event noise~\cite{red_xu2021motion,zhang2022unifying,sun2022event,shang2021bringing,teng2022nest,chen2024motion,xu2025motion,cho2023non,kim2023event,kim2024frequency}. 
\re{Specifically, Shang \etal~\cite{shang2021bringing} propose D2Net, a framework designed for handling non-consecutive blurry video frames. Semi-supervised~\cite{red_xu2021motion} and self-supervised~\cite{zhang2023generalizing} approaches utilize real-world events to mitigate performance degradation resulting from data inconsistency. The two-stage restoration network EF-Net~\cite{sun2022event} incorporates a cross-modal attention module and a symmetric cumulative event representation to integrate event and image information effectively. More recently, a motion-adaptive transformer~\cite{xu2025motion} has been introduced to model non-local dependencies by jointly leveraging intensity and spatial cues. A frequency-aware event-based model~\cite{kim2024frequency} has also been proposed to actively exploit temporal frequency information for improved deblurring performance.}
\par
\re{Although a few recent works attempt to jointly estimate deblurring and interpolation~\cite{flawless_jin2019learning,bin_shen2020blurry,zhong2023blur}, the majority of deblurring methods still focus solely on restoring latent sharp images aligned with the blurry frame’s exposure time, without explicitly reconstructing intermediate latent frames across blurry inputs.} 

\subsection{Frame Interpolation}\label{subSec:Interpolation}
Existing frame-based interpolation methods can be roughly categorized into warping-based and kernel-based approaches. Warping-based approaches \cite{superslomo_jiang2018super,softmax_niklaus2020softmax,44_xue2019video,dain_bao2019depth} generally combine optical flow \cite{8flownet_ilg2017flownet,PWC_sun2018pwc} with image warping to predict intermediate frames and several techniques have been proposed to enhance the interpolation performance, \eg, forward warping \cite{softmax_niklaus2020softmax}, spatial transformer networks \cite{44_xue2019video}, and depth information \cite{dain_bao2019depth}. However, these methods often assume linear motion and brightness constancy between two reference frames, thus failing to handle arbitrary motions.
Rather than warping reference frames with optical flow, kernel-based methods \cite{23AdaConv_niklaus2017video} model the frame interpolation as a local convolution on the reference frames, where the kernel is directly estimated from the input frames. Although kernel-based methods are more robust to complex motions and brightness changes, their scalability is often limited by the fixed sizes of convolution kernels.

\par
The common challenge of frame-based interpolation is the missing information between reference frames, which can be alleviated by leveraging the extremely low latency of events. Recent advancements~\cite{timelens_tulyakov2021time,tulyakov2022time,he2022timereplayer,yang2024latency,cho2024tta} in this field have harnessed the complementary strengths of both frames and events, resulting in exceptional interpolation results even in scenarios involving non-linear motions. \re{TimeLens~\cite{timelens_tulyakov2021time} pioneers a unified interpolation framework by combining warping-based and synthesis-based approaches. Building on this foundation, TimeLens++~\cite{tulyakov2022time} introduces a method to estimate reliable non-linear continuous flow from sparse event inputs. TimeReplayer~\cite{he2022timereplayer} presents an unsupervised framework that facilitates the video interpolation process. More recently, Cho \etal~\cite{cho2024tta} propose a test-time adaptation framework for video interpolation, incorporating patch-mixed sampling to mitigate overfitting. However, these methods usually require high-quality reference frames for interpolation, and thus often struggle to recover sharp results when encountering motion blur.}



\subsection{Joint Deblurring and Interpolation}\label{subSec:Joint}
Previous frame-based methods have approached the joint deblurring and interpolation task \cite{flawless_jin2019learning,bin_shen2020blurry,zhong2023blur}. The work of \cite{flawless_jin2019learning} presents a cascaded scheme that first deblurs and then interpolates. To mitigate the accumulated errors introduced in the cascaded scheme, Shen \etal \cite{bin_shen2020blurry} propose a pyramid recurrent framework \cite{bin_shen2020blurry} to estimate the latent sharp sequence. BiT~\cite{zhong2023blur} designs a transformer-based model for arbitrary time motion from blur reconstruction. \re{Recently, many event-based methods~\cite{ledvdi_lin2020learning,weng2023event,sun2023event,yang2024latency,sun2024unified} demonstrate impressive results by designing an effective aggregation module of event streams and frames. LEDVDI~\cite{ledvdi_lin2020learning} is categorized into the cascaded scheme as it achieves deblurring and interpolation with different stages, while frameworks proposed in~\cite{weng2023event,sun2023event,yang2024latency,sun2024unified} perform deblurring and interpolation with a one-stage model. However, all the aforementioned methods require supervised training on synthetic datasets, which limits their performance in real-world scenarios due to data inconsistency. In contrast, our EVDI++ fulfills joint deblurring and interpolation with a self-supervised learning framework, directly fitting real-world data distributions.}

\section{Problem Formulation}\label{Sec:ProbState}
Videoing highly dynamic scenes often suffers from blurry artifacts and the absence of inter-frame information; therefore, Video Enhancement (VE) plays a crucial role in visual perception. Given two consecutive reference frames $F_i, F_{i+1}$ captured within the exposure time $\mathcal{T}_i, \mathcal{T}_{i+1}$ and the corresponding event streams $\mathcal{E}_{i+1}^{i}$ triggered inside $\mathcal{T}_{i+1}^{i}$, where $\mathcal{T}_{i+1}^{i}\triangleq \mathcal{T}_i\cup \mathcal{T}_{i\rightarrow i+1}\cup \mathcal{T}_{i+1}$ with $\mathcal{T}_{i\rightarrow i+1}$ indicating the time interval between $F_i$ and $F_{i+1}$, the task of EVDI++ is to achieve VE directly from reference frames, \ie,
\begin{equation}\label{ps_EVDI++}
    L(t) = \operatorname{EVDI\text{++}}(t; F_i, F_{i+1},\mathcal{E}^{i}_{i+1}),\quad \forall t\in \mathcal{T}^{i}_{i+1},
\end{equation}
where $L(t)$ indicates the latent image of arbitrary time
$t\in \mathcal{T}^{i}_{i+1}$, $F\in\{B,I\}$, and $B$ and $I$ denote blurry image and sharp image, respectively. 
According to \cref{ps_EVDI++}, EVDI++ degrades to (i) Motion Deblurring (MD) when $t\in \{\mathcal{T}_i, \mathcal{T}_{i+1}\}$ and $F=B$; (ii) Frame Interpolation (FI) when $t\in \mathcal{T}_{i \rightarrow i+1}$ and $F=I$; (iii) Blurry Video Enhancement (BVE) when $t\in \mathcal{T}_{i+1}^{i}$ and $F=B$. 
Thus, EVDI++ is more general than MD, FI, and BVE, and provides a unified formulation for the VE task.

\par 
\noindent \textbf{{\myname} \vs ~Motion Deblurring.} The MD aims at reconstructing the sharp latent images $\{L(t)\}_{t\in \mathcal{T}_i}$ from the corresponding blurry frame $B_i$. Providing the concurrent event streams $\mathcal{E}_i$ triggered within $\mathcal{T}_i$, we have
\begin{equation}\label{Deblur}
    L(t) = \operatorname{MD}(t; B_i, \mathcal{E}_i),\quad t\in \mathcal{T}_i,
\end{equation}
Existing deblurring methods \cite{Deng_2021_ICCV,Suin_2021_CVPR} focus mainly on recovering latent frames inside the exposure time $\mathcal{T}_i$, while \myname\ can predict the latent images of the time instance both inside the exposure time $\mathcal{T}_i$ (or $\mathcal{T}_{i+1}$) and between blurry frames $\mathcal{T}_{i \rightarrow i+1}$, as shown in Fig.~\ref{Overview}. 

\par 
\noindent \textbf{{\myname} \vs ~Frame Interpolation.} Conventional FI task aims at recovering the intermediate latent images $\{L(t)\}_{t\in \mathcal{T}_{i \rightarrow i+1}}$ from sharp reference frames $I_i,\ I_{i+1}$. Providing the concurrent event streams $\mathcal{E}_{i \rightarrow i+1}$ emitted within $\mathcal{T}_{i\rightarrow i+1}$, we have
\begin{equation}\label{Interp}
    L(t) = \operatorname{FI}(t; I_i, I_{i+1},\mathcal{E}_{i \rightarrow i+1}),\quad t\in \mathcal{T}_{i \rightarrow i+1},
\end{equation}
Most FI methods \cite{superslomo_jiang2018super,dain_bao2019depth,timelens_tulyakov2021time} are designed to restore inter-frame latent images from high-quality (sharp and clear) reference frames $I_i,\ I_{i+1}$, while \myname\ can directly accept both sharp or blurry inputs, which is more challenging than conventional FI.

\noindent\textbf{{\myname} \vs ~Blurry Video Enhancement.} The BVE techniques aim at recovering sharp latent images $\{L(t)\}_{t\in \mathcal{T}^{i}_{i+1}}$ from blurry frames $B_i$, $B_{i+1}$. Providing the concurrent event streams $\mathcal{E}^{i}_{i+1}$ emitted within $\mathcal{T}^{i}_{i+1}$, we have
\begin{equation}\label{BlurryVE}
    L(t) = \operatorname{BVE}(t; B_i, B_{i+1},\mathcal{E}^{i}_{i+1}),\quad t\in \mathcal{T}^{i}_{i+1}.
\end{equation}
Most BVE methods \cite{ledvdi_lin2020learning,weng2023event} predict latent images of time instances both within the exposure time $\mathcal{T}_i$ (or $\mathcal{T}_{i+1}$) and between blurry frames $\mathcal{T}_{i \rightarrow i+1}$. However, these methods tend to suffer from accumulated errors in cascaded schemes, resulting in sub-optimal performance. Additionally, they require supervised training on synthetic datasets, which limits their performance in real-world scenarios due to inconsistencies in the data. 

Ideally, \myname\ can approach the VE task by unifying MD, FI, and BVE in \cref{ps_EVDI++}. However, challenges still exist to efficiently realize \myname\ in real-world scenarios.

\begin{itemize}
    \item MD, FI, and BVE should be simultaneously addressed in a unified framework to fulfill the \myname. Most previous attempts for VE \cite{flawless_jin2019learning, ledvdi_lin2020learning} employ a cascaded scheme that performs frame interpolation after deblurring, which often propagates deblurring error to the interpolation stage and thus leads to sub-optimal results.
    \item Existing related methods are generally developed within a supervised learning framework \cite{ledvdi_lin2020learning,bin_shen2020blurry,weng2023event,sun2023event}, of which supervision is usually provided by synthetic blurry images and events. Hence, the performance might degrade in real scenes due to the different distribution between synthetic and real-world data.
\end{itemize}

Our previous work, \ie, EVDI~\cite{zhang2022unifying}, propose a self-supervised framework to reconstruct sharp latent images $\{L(t)\}_{t\in \mathcal{T}^{i}_{i+1}}$ from blurry frames $B_i$, $B_{i+1}$ and concurrent event streams $\mathcal{E}^{i}_{i+1}$ emitted within $\mathcal{T}^{i}_{i+1}$, which can be formulated as
\begin{equation}\label{pf_EVDI}
    L(t) = \operatorname{EVDI}(t; B_i, B_{i+1},\mathcal{E}^{i}_{i+1}),\quad \forall t\in \mathcal{T}^{i}_{i+1}.
\end{equation}
Comparing \cref{ps_EVDI++} and \cref{pf_EVDI} reveals that our proposed EVDI++ is more general than its previous version, \ie, EVDI, since EVDI++ accepts both blurry and sharp images, whereas EVDI is specialized for blurry inputs.

\begin{figure*}[!t]
  \centering
    \includegraphics[width=0.90\linewidth]{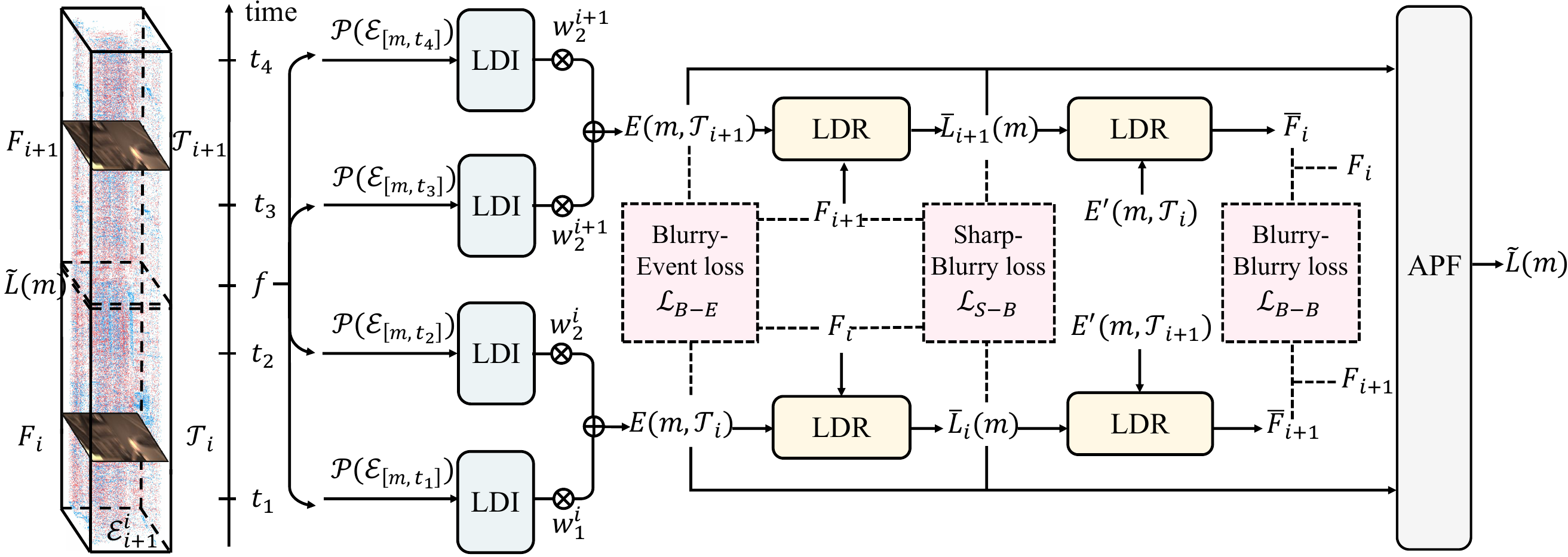}
    \caption{Architecture of the proposed EVDI++, which is composed of a Learnable Double Integral (LDI) module, a Learning-based Division Reconstruction (LDR) module, and an Adaptive Parameter-free Fusion (APF) module.}
    \label{framework}
    \vspace{-3mm}
\end{figure*}

\section{Method}
We first analyze the feasibility of simultaneous deblurring and interpolation by revisiting the physical model and theoretical derivation in \cref{Subsec:Feasibility} and then introduce the Learnable Double Integral (LDI) network, a pivotal element shared by our previous work, \ie, EVDI~\cite{zhang2022unifying}, and the proposed EVDI++ in \cref{NetworkArchi}. A critical reassessment of the EVDI structure and its limitations is provided in \cref{EVDI}. Finally, we present EVDI++, which features a Learning-based Division Reconstruction (LDR) module, an Adaptive Parameter-free Fusion (APF) module, and corresponding self-supervised losses, as shown in \cref{EVDI++}. This effectively addresses the identified limitations in EVDI. The overall data flow of event pre-processing and LDI is illustrated in \cref{framework} (a), and the data flows of the reconstruction and self-supervised framework in EVDI and EVDI++ are depicted in \cref{framework} (b) and (c), respectively. 

\subsection{Feasibility Analysis}\label{Subsec:Feasibility}
To analyze the feasibility of simultaneously fulfilling motion deblurring, frame interpolation, and blurry video enhancement, we first establish the relationship among the sharp latent images $L(m)$ at arbitrary timestamps $m$, the reference image $F\in\{B, I\}$, and corresponding event streams $\mathcal{E}$ by defining a mapping function, \ie, 
\begin{equation}\label{mapping_function}
    L(m) = \Phi(m;F,\mathcal{E}).
\end{equation}
By utilizing $F\in\{B,I\}$ and $\mathcal{E}$ as inputs, we employ a unified mapping function $\Phi(\cdot)$ to estimate latent sharp frames with arbitrary timestamps, thereby accomplishing three tasks concurrently. Then, we analyze how this unified mapping function is constructed for performing the reconstruction tasks associated with each objective.

\noindent \textbf{Motion Deblurring.} \cref{mapping_function} represents MD when $F=B$ and $m\in{\mathcal{T}}$ with $\mathcal{T}=[t_s,t_s+T]$ representing the exposure interval of the reference image from the start timestamps $t_s$ to the end timestamps $t_s+T$. The specific expression of the mapping function $\Phi(\cdot)$ can be obtained by revisiting the physical generation model of the blurry frames and event streams. On the one hand, events $\mathcal{E}\triangleq \{(\mathbf{x}_n,t_n,p_n)\}_{t_n\in \mathcal{T}}$ are triggered whenever the log-scale brightness change exceeds the event threshold $c>0$, \ie,
\begin{equation}\label{latent_expEvent}
    \operatorname{log}(L(t_n,\mathbf{x}_n)) - \operatorname{log}(L(t_{n}+\Delta{t},\mathbf{x}_n)) = p_n\cdot c,
\end{equation}
where $L(t_n,\mathbf{x}_n)$ and $L(t_n+\Delta{t},\mathbf{x}_n)$ denote the instantaneous intensity at time $t_n$ and $t_n+\Delta{t}$ at the pixel position $\mathbf{x}_n$, and polarity $p_n\in\{+1,-1\}$ indicates the direction of brightness changes. On the other hand, blurry images can be formulated as the average of latent images within the exposure time \cite{chen2018reblur2deblur}, \ie,
\begin{equation}\label{blurry_latent}
    B = \frac{1}{T} \int_{t\in \mathcal{T}} L(t) dt.
\end{equation}
Combining \cref{latent_expEvent} and \cref{blurry_latent}, and the EDI model~\cite{edi_pan2019bringing}, the expression of $\Phi(\cdot)$ can be represented as
\begin{align}
     L(m) &= \Phi(m;F,\mathcal{E}_\mathcal{T})=  \frac{F}{E(m,\mathcal{T})} , \quad \text{with} \label{blurry_latent_event} 
    \\
    E(m,\mathcal{T}) &= \frac{1}{T} \int_{t \in \mathcal{T}} \operatorname{exp} (c\int_m^t e(s)ds)dt \label{EDI} 
\end{align}
representing the relation between reference frames $F$ and latent sharp images $L(m)$ from the perspective events $\mathcal{E}_\mathcal{T}$. Moreover, we omit the pixel position in \cref{blurry_latent,EDI} for better readability. In \cref{EDI}, $e(t)\triangleq \sum_{i} p_i \cdot \delta(t-t_i)$ denotes the continuous representation of events with $\delta(\cdot)$ indicating the Dirac function. 

\noindent \textbf{Frame Interpolation.}
\cref{mapping_function} denotes FI when $F=I$ that represents $T\rightarrow0$ and the expression of $\Phi(\cdot)$ is the same as as \cref{blurry_latent_event}, which constitute a unique case within the double integral of $E(\cdot)$ as  
\begin{equation}
\begin{aligned}\label{EDI_sharp}
    E(m,\mathcal{T})&=\lim\limits_{T \to 0} \frac{\int_{t_s}^{t_s+T} \operatorname{exp} (c\int_m^t e(s)ds)dt}{T},
    \\
     &= \operatorname{exp} (c\int_m^{t_s} e(s)ds).
\end{aligned}
\end{equation}

\begin{figure}[!t]
  \centering
    \includegraphics[width=0.82\linewidth]{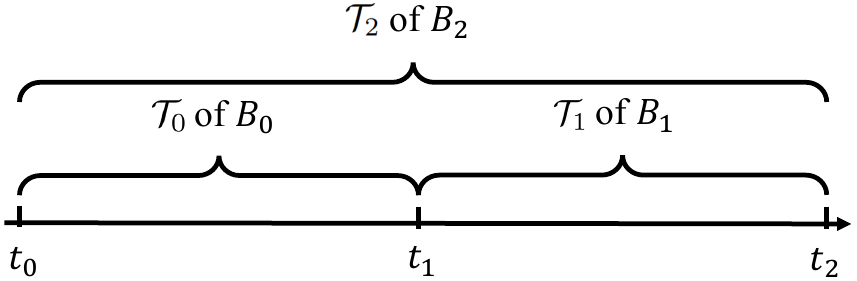}
    \vspace{-2mm}
    \caption{An illustration of three blurry frames, where the target timestamp $t_0$ falls within the exposure time of $B_{1}$ and $B_{2}$, but lies outside the exposure time of $B_{1}$.}
    \label{FA}
    \vspace{-3mm}
\end{figure}

\noindent \textbf{Blurry Video Enhancement.}
\cref{mapping_function} means BVE when $F=B$ and $m\notin \mathcal{T}$, and the expression of $\Phi(\cdot)$ can also be represented as \cref{blurry_latent_event}. Moreover, \cref{blurry_latent_event} can also be extended to recover the latent frames when $m\notin \mathcal{T}$ to fulfill BVE. To prove this, we assume a target arbitrary timestamps $t_0$ that lies outside of blurry frame $B_1$ with exposure interval $\mathcal{T}_1=[t_1,t_2]$, as shown in \cref{FA}. Subsequently, two blurry frames $B_0$ and $B_2$ can be constructed, with their respective exposure intervals being $\mathcal{T}_0=[t_0,t_1]$ and $\mathcal{T}_0=[t_0,t_2]$. Since $t_0$ locates inside both $\mathcal{T}_{0}$ and $\mathcal{T}_{2}$, we can directly derive the following equations using \cref{blurry_latent_event}:
\begin{equation}
\begin{aligned}
    B_{0} &= L(t_0)\cdot E(t_0, \mathcal{T}_{0}),
    \\
    B_{2} &= L(t_0)\cdot E(t_0, \mathcal{T}_{2}).
\label{BLE}
\end{aligned}
\end{equation}
The left and right sides of the first equation in \cref{BLE} are multiplied by $T_0$, while the left and right sides of the second equation in \cref{BLE} are multiplied by $T_2$. The resulting equations are then subtracted from each other; therefore, one can obtain
\begin{equation}
    T_{2}B_{2} - T_{0}B_{0} = L(t_0) \cdot (T_{2}E(t_0, \mathcal{T}_{2}) - T_{0}E(t_0, \mathcal{T}_{0})).
\label{E13}
\end{equation}
Based on \cref{blurry_latent} and \cref{EDI}, we also have
\begin{equation}
\begin{aligned}
    T_{1}B_{1} &= T_{2}B_{2} - T_{0}B_{0},
    \\
    T_{1}E(t_0,\mathcal{T}_{1}) &= T_{2}E(t_0, \mathcal{T}_{2}) - T_{0}E(t_0, \mathcal{T}_{0}).
\label{E14}
\end{aligned}
\end{equation}
Combining \cref{E13} and \cref{E14} can easily derive 
\begin{equation}\label{LBE}
    L(t_0) = \frac{B_{1}}{E(t_0, \mathcal{T}_{1})},
\end{equation}
which means \cref{blurry_latent_event} can also be applied to recover the latent images, \eg, $L(t_0)$, located outside the exposure period of the blurry frames, \eg, $B_{1}$. 

Therefore, leveraging \cref{blurry_latent_event,EDI} for the reference image $F\in\{B,I\}$ enables a unified approach for MD, FI, and BVE at arbitrary timestamps in theory. Nevertheless, employing \cref{blurry_latent_event,EDI} directly for unified deblurring and interpolation by estimating the event integral $E(\cdot)$ faces several challenges: Firstly, the computation of $E(\cdot)$ requires the knowledge of the event threshold $c$, which significantly impacts the recovery performance~\cite{edi_pan2019bringing} but is challenging to precisely estimate due to its temporal instability~\cite{survey_9138762}. Secondly, real-world events are noisy due to the non-ideality of physical sensors~\cite{survey_9138762}, \eg, limited read-out bandwidth, and thus often lead to degraded results, especially when encountering long-term integrals of events where $E(\cdot)$ is severely contaminated by noises. Hence, we propose to employ a learning-based architecture to fit the statistics of real-world events.


\begin{figure}[!t]
  \centering
    \includegraphics[width=0.82\linewidth]{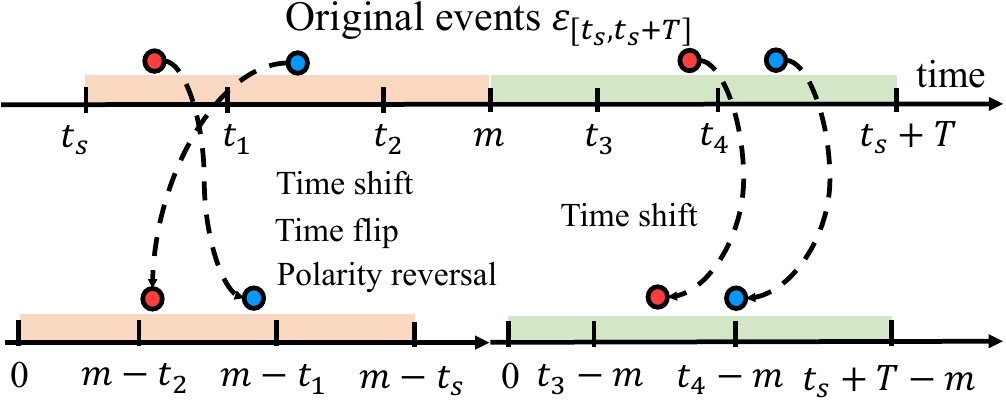}
    \vspace{-2mm}
    \caption{Example of the pre-processing operation. The left event subset experiences time shift, flip, and polarity reversal, \ie, $\mathcal{R}(\cdot)$, since $t_s-m<0$, and the right subset is processed by the operator of time shift, \ie, $\mathcal{S}(\cdot)$, as $t_s+T-m\geq 0$.}
    \label{PP}
    \vspace{-3mm}
\end{figure}

\subsection{Learnable Double Integral Network}\label{NetworkArchi}
As shown in \cref{framework}, the Learnable Double Integral (LDI) network receives event streams $\mathcal{E}_{i+1}^{i}$ as input and outputs the event double integral $E(m,\mathcal{T}_i)$ and $E(m,\mathcal{T}_{i+1})$ for EVDI~\cite{zhang2022unifying} and EVDI++. Suppose the LDI for double integral is trained to approximate a specific case $E(0,\mathcal{T}_{[0,T]}) \approx \operatorname{LDI}(\mathcal{E}_{[0,T]})$ \ie,
\begin{equation}\label{LDI}
    \operatorname{LDI}(\mathcal{E}_{[0,T]}) \approx \frac{1}{T} \int_0^T \operatorname{exp}(c \int_0^t e(s) ds) dt,
\end{equation}
where $\mathcal{T}_{[0,T]}$ indicates the time interval from $0$ to $T>0$ and $\mathcal{E}_{[0,T]}$ is the corresponding event streams.
Now we consider a more general case of $E(m,\mathcal{T})$, which can be written as
\begin{equation}\label{EDI_divide}
\begin{aligned}
     E(m,\mathcal{T}) =& \frac{1}{T}  \int_{t_s}^{m} \operatorname{exp} (c\int_m^t e(s)ds)dt 
     \\
     &+ \frac{1}{T} \int_{m}^{t_s+T} \operatorname{exp} (c\int_m^t e(s)ds)dt,
\end{aligned}
\end{equation}
where $t_s$ indicates the starting time of $\mathcal{T}$. Applying $t'=t-m$ and $s'=s-m$ to Eq.~\eqref{EDI_divide}, we have 
\begin{equation}\label{E_2G}
\begin{aligned}
    E(m,\mathcal{T}) =& -\frac{1}{T} \int_0^{t_s - m} \operatorname{exp}(c\int_0^{t'} e(s'+m)ds')dt'
    \\
    &+ \frac{1}{T} \int_0^{t_s+T-m} \operatorname{exp}(c\int_0^{t'} e(s'+m)ds')dt'
    \\
    =& w_1 G(\mathcal{E}_{[m,t_s]}) + w_2 G(\mathcal{E}_{[m,t_s+T]}),
\end{aligned}
\end{equation}
where $w_1=(m-t_s)/T,\ w_2=(t_s+T-m)/T$ are weights and $G(\cdot)$ is a general formula defined as
\begin{equation}\label{G}
    G(\mathcal{E}_{[m,t_r]}) = \frac{1}{t_r-m}\int_0^{t_r-m} \operatorname{exp}(c\int_0^t e(s+m) ds) dt
\end{equation}
and $t_r$ denotes the reference time. By comparing \cref{LDI} and \cref{G}, it becomes evident that they exhibit significant similarity under the condition that $t_r-m=T$, with the sole distinction being between $e(s)$ and $e(s+m)$. If \cref{LDI} can be adjusted to match \cref{G}, then LDI can be used to approximate the general event double integral. Consequently, we propose a series of operations, including time shifting, temporal inversion, and polarity reversal, to facilitate the alignment of \cref{LDI} with \cref{G}. Based on the above definition, we can calculate $E(m,\mathcal{T})$ by approximating \cref{G} with the LDI, \ie, \cref{LDI}. For the case of $t_r-m\geq 0$, $G(\cdot)$ can be directly approximated by 
\begin{equation}
    G(\mathcal{E}_{[m,t_r]}) \approx \operatorname{LDI}(\mathcal{S}(\mathcal{E}_{[m,t_r]}))
\end{equation}
with $\mathcal{S}(\mathcal{E}_{[m,t_r]}) = \{(\mathbf{x},t-m,p) | (\mathbf{x},t,p)\in \mathcal{E},t\in [m,t_r]\}$ representing the event operator of time shift. For the case of $t_r-m< 0$, 
\begin{equation}
\begin{aligned}
    G(\mathcal{E}_{[m,t_r]}) &= \frac{1}{m-t_r} \int_0^{m-t_r} \operatorname{exp}(c\int_0^t -e(-s+m) ds)dt
    \\ 
    &\approx \operatorname{LDI}(\mathcal{R}(\mathcal{E}_{[m,t_r]})),
\end{aligned}
\end{equation}
where $\mathcal{R}(\mathcal{E}_{[m,t_r]}) = \{(\mathbf{x},-t+m,-p) | (\mathbf{x},t,p)\in \mathcal{E},t\in [t_r,m]\}$ indicates the event operator composed of time shift, flip and polarity reversal, as shown in \cref{PP}. For simplicity, we define a unified pre-processing operator $\mathcal{P}(\cdot)$ as follows.
\begin{equation}\label{P}
    \mathcal{P}(\mathcal{E}_{[m,t_r]}) =\left\{
	\begin{array}{ll}
	\mathcal{S}(\mathcal{E}_{[m,t_r]}) & \text { if } t_r \geq m, \\
	\mathcal{R}(\mathcal{E}_{[m,t_r]}) & \text { if } t_r < m.
	\end{array}\right.
\end{equation}
Thus, Eq.~\eqref{E_2G} can be reformulated as
\begin{equation}\label{LDI_E}
    E(m,\mathcal{T}) \approx w_1 \operatorname{LDI}(\mathcal{P}(\mathcal{E}_{[m,t_s]})) + w_2 \operatorname{LDI}(\mathcal{P}(\mathcal{E}_{[m,t_s+T]})),
\end{equation}
meaning that arbitrary $E(m,\mathcal{T})$ can be approximated by a weighted combination of LDI outputs, where the LDI only needs to be trained once to fit the case of Eq.~\eqref{LDI}. 

\begin{figure}[!t]
  \centering
   \includegraphics[width=0.99\linewidth]{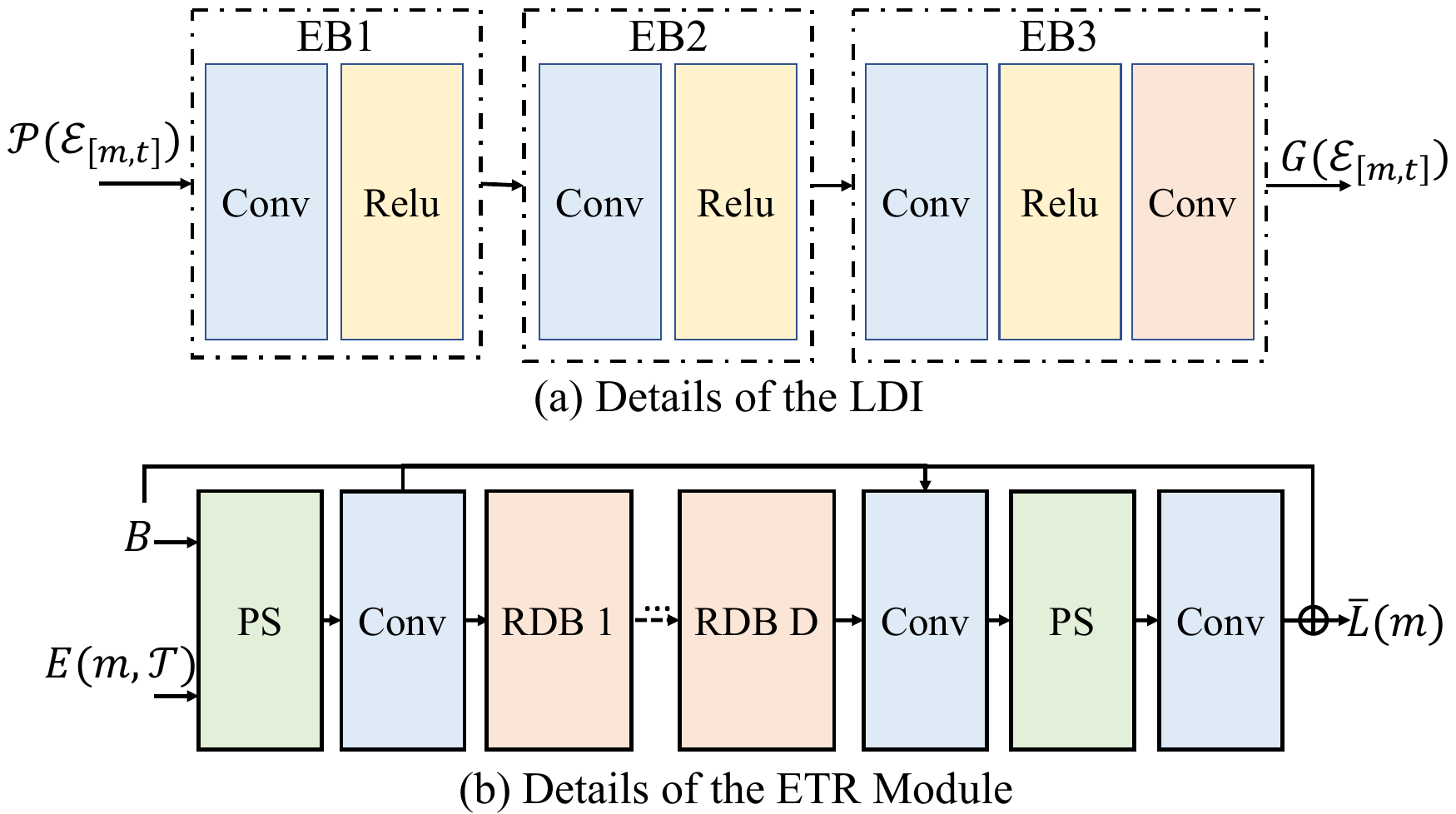}
    \vspace{-0mm}
   \caption{(a) The detailed architecture of our LDI consists of three Encode Blocks (EB). (b) An illustration of the Exposure-Transferred Reconstruction (ETR) Module that consists of multiple Residual Dense Blocks (RDBs), and Pixel Shuffle (PS) modules.}
   \label{fig:LDI_framework}
   \vspace{-3mm}
\end{figure}

\par 
To this end, as shown in \cref{fig:LDI_framework} (a), we design a simple yet efficient architecture to achieve an LDI network, which consists of three encode blocks, each utilizing $3\times3$ convolutional layers with ReLU activation functions. The convolutional layers are configured with output channels of 32, 64, 32, and 1. For the input of the LDI, we introduce a spatio-temporal event representation. With a pre-defined number, \eg, $N$, we fairly divide $N$ temporal bins from $t=0$ to $t=T_{i+1}^{i}$ where $T_{i+1}^{i}$ denotes the total duration of $\mathcal{T}_{i+1}^{i}$. We then accumulate the events pre-processed by $\mathcal{P}(\cdot)$ inside each temporal bin and form a $2N \times H \times W$ tensor as the LDI input with $2, H, W$ indicating event polarity, image height, and width, respectively. Therefore, our event representation enables flexible choice of the target timestamp $m$ while maintaining a fixed input format, which allows the network to restore the latent images $L(m)$ at arbitrary $m\in \mathcal{T}_{i+1}^{i}$ without any network modification or re-training process.

\def\imwidth{0.33}
\def\cimwid{0.14}

\def\zuoxia{(-0.3,-0.9)}
\def\youshang{(0.2,0.45)}

\def\ssyy{(-0.8,-0.85)}
\def\ssizzone{1.2cm}
\def\sswidth{0.245\textwidth}
\def\ssmag{4}
\def\scc{(1.43,-0.95)}

\def\ssxxsone{(0.65, -0.00)}
\def\ssyysone{(-0.22, -0.45)}

\def\ssxxstwo{(0.70, 0.05)}
\def\ssyystwo{(-0.22, -0.45)}

\def\ssxxsthree{(0.87, 0.20)}
\def\ssyysthree{(-0.22, -0.45)}

\newcommand{\newimgfontt}{\fontsize{7}{5}\selectfont}

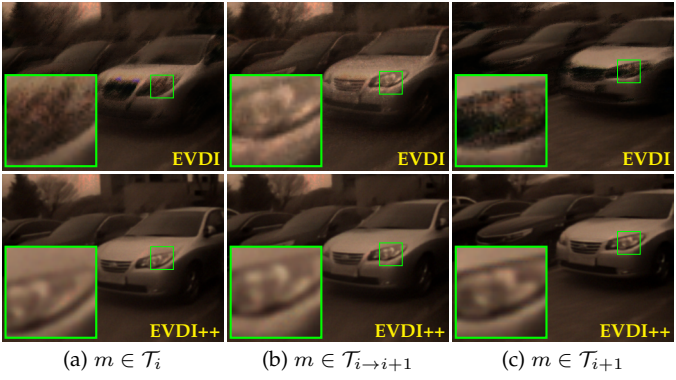
\begin{figure}[!htb]
\footnotesize
	\centering
\begin{minipage}[t]{\imwidth\linewidth}
    		\centering
			\begin{tikzpicture}[spy using outlines={rectangle,green,magnification=\ssmag,size=\ssizzone},inner sep=0]
				\node {\includegraphics[width=\linewidth]{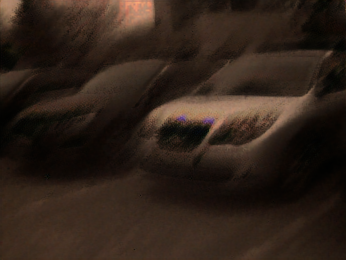}};
                \spy on \ssxxsone in node [left] at \ssyysone;
                \node [anchor=east, font=\newimgfontt] at \scc {\textcolor{yellow}{\bf EVDI}};
				\end{tikzpicture} \\
            \vspace{.2em}
            \centering
    	    \begin{tikzpicture}[spy using outlines={green,magnification=\ssmag,size=\ssizzone},inner sep=0]
				\node {\includegraphics[width=\linewidth]{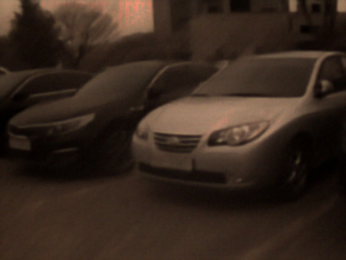}};
                \spy on \ssxxsone in node [left] at \ssyysone;
                \node [anchor=east, font=\newimgfontt] at \scc {\textcolor{yellow}{\bf EVDI++}};
				\end{tikzpicture}
            (a) $m\in \mathcal{T}_{i}$ 
\end{minipage}%
    \hfill
\begin{minipage}[t]{\imwidth\linewidth}
    		\centering
    	    \begin{tikzpicture}[spy using outlines={green,magnification=\ssmag,size=\ssizzone},inner sep=0]
				\node {\includegraphics[width=\linewidth]{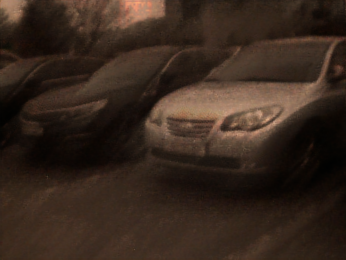}};
                \spy on \ssxxstwo in node [left] at \ssyystwo;
                \node [anchor=east, font=\newimgfontt] at \scc {\textcolor{yellow}{\bf EVDI}};
				\end{tikzpicture}
            \\
            \centering
            \vspace{.2em}
    	    \begin{tikzpicture}[spy using outlines={green,magnification=\ssmag,size=\ssizzone},inner sep=0]
				\node {\includegraphics[width=\linewidth]{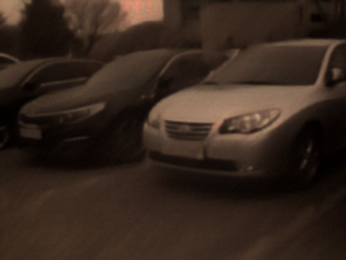}};
                \spy on \ssxxstwo in node [left] at \ssyystwo;
                \node [anchor=east, font=\newimgfontt] at \scc {\textcolor{yellow}{\bf EVDI++}};
				\end{tikzpicture}
            (b) $m\in \mathcal{T}_{i\rightarrow {i+1}}$
\end{minipage}%
    \hfill
\begin{minipage}[t]{\imwidth\linewidth}
    		\centering
    	    \begin{tikzpicture}[spy using outlines={green,magnification=\ssmag,size=\ssizzone},inner sep=0]
				\node {\includegraphics[width=\linewidth]{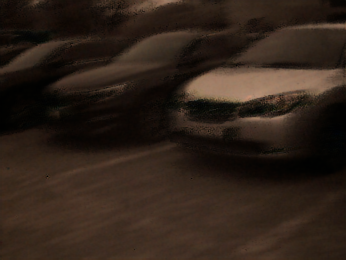}};
                \spy on \ssxxsthree in node [left] at \ssyysthree;
                \node [anchor=east, font=\newimgfontt] at \scc {\textcolor{yellow}{\bf EVDI}};
				\end{tikzpicture} \\
            \centering
            \vspace{.2em}
    	    \begin{tikzpicture}[spy using outlines={green,magnification=\ssmag,size=\ssizzone},inner sep=0]
				\node {\includegraphics[width=\linewidth]{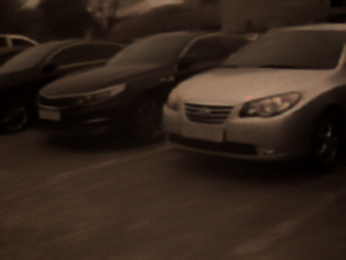}};
                \spy on \ssxxsthree in node [left] at \ssyysthree;
                \node [anchor=east, font=\newimgfontt] at \scc {\textcolor{yellow}{\bf EVDI++}};
				\end{tikzpicture}
                (c) $m\in \mathcal{T}_{i+1}$
    	\end{minipage}%

\caption{(a), (b), and (c) denote the reconstructed latent sharp images $\tilde{L}(m)$ by using EVDI (first row) and EVDI++ (second row) with $m\in \mathcal{T}_{i}$, $m\in \mathcal{T}_{i\rightarrow {i+1}}$, and $m\in \mathcal{T}_{i+1}$, respectively.}
  \vspace{-1em}
    \label{EVDI_fusion}
  \end{figure}

The LDI not only consolidates the tasks of Motion Deblurring (MD) and Blurry Video Enhancement (BVE) by leveraging their same input of blurry reference frames ($B_i$ and $B_{i+1}$), but also addresses the issue of Frame Interpolation (FI) when the input transitions from blurry to sharp reference images ($I_i$ and $I_{i+1}$). This transition leads to the degeneration of the exposure period duration $T$ to zero, and thus \cref{LDI_E} can be rewritten as
\begin{equation}
\begin{aligned}\label{E_1G}
    E(m,\mathcal{T}) &\approx \eta_1\operatorname{LDI}(\mathcal{P}(\mathcal{E}_{[m,t_s]})) + \eta_2 \operatorname{LDI}(\mathcal{P}(\mathcal{E}_{[m,t_s+0]})),
    \\
     &\approx (\eta_1+\eta_2)\operatorname{LDI}(\mathcal{P}(\mathcal{E}_{[m,t_s]})).
\end{aligned}
\end{equation}
Hence, we can seamlessly implement FI by using two identical LDI to approximate \cref{E_1G} with $\eta_1+\eta_2=1$, without requiring any detailed modifications to the proposed architecture. 


\subsection{EVDI}\label{EVDI}
We briefly review our previous work, EVDI~\cite{zhang2022unifying}, and analyze its limitations, focusing primarily on two aspects beyond the LDI module: the reconstruction network and the self-supervised losses.

\subsubsection{Reconstruction Network} 
\par
\noindent \textbf{Fusion Network with learnable parameters.}
After obtaining $E(m,\mathcal{T})$ from the LDI, the latent image $L(m)$ can be coarsely restored by Eq.~\eqref{blurry_latent_event}. The latent images reconstructed from $B_i, B_{i+1}$ can be denoted as $L_i(m), L_{i+1}(m)$, respectively, and manually generate an extra result $L^{i}_{i+1}(m)$ by
\begin{equation}
L^{i}_{i+1}(m) = \omega(m) L_i(m) + (1-\omega(m)) L_{i+1}(m),
\end{equation}
where $\omega(m)$ denotes the weighting function which can be viewed in Eq.~(18) of \cite{zhang2022unifying}. Finally, the fusion network receives $L_{i}(m)$, $L_{i+1}(m)$, $L_{i+1}^{i}(m)$, $E(m,\mathcal{T}_{i})$, $E(m,\mathcal{T}_{i+1})$ and produces the final latent image $\tilde{L}(m)$, which consists of convolutional, residual, and spatial and channel attention blocks with learnable parameters.

\par
\noindent \textbf{Limitations.} 
The fusion network with learnable parameters, taking multiple inputs ($L_{i}(m)$, $L_{i+1}(m)$, $L_{i+1}^{i}(m)$, $E(m,\mathcal{T}_{i})$, and $E(m,\mathcal{T}_{i+1})$), exhibits two limitations in terms of efficiency and robustness. On the one hand, the learnable parameters, \ie, self-attention blocks, increase the complexity of the model, thus compromising overall efficiency. On the other hand, the fusion network is beneficial for reconstruction within the exposure interval $\mathcal{T}_{i\rightarrow {i+1}}$, illustrated in \cref{EVDI_fusion} (c). However, reconstructing images within the exposure interval $\mathcal{T}_{i}$ and $\mathcal{T}_{i+1}$ with extended time intervals based on $B_{i+1}$ and $B_{i}$ yields less reliable results,  as depicted in \cref{EVDI_fusion} (a) and (b), introducing extra artifacts and consequently degrading overall fusion quality.

\subsubsection{Self-supervised Learning}
The self-supervised learning framework in EVDI~\cite{zhang2022unifying} consists of three different losses, each formulated based on mutual constraints among blurry frames, sharp latent images, and event streams.

\par 
\noindent \textbf{Blurry-Event Loss.} 
The double integral of events $E(m,\mathcal{T})$ corresponds to the mapping relation between blurry frames and sharp latent images. For multiple blurry inputs, the EVDI formulates the consistency between the latent images reconstructed from different blurry frames, \eg, $L_i(m) = L_{i+1}(m)$. Considering the quantization error which might be accumulated in $E(m,\mathcal{T})$, we rewrite the consistency as
\begin{equation}\label{E2B2-div}
    \frac{B_i}{E(m,\mathcal{T}_i)} \approx \frac{B_{i+1}}{E(m,\mathcal{T}_{i+1})},
\end{equation}
where $E(m,\mathcal{T}_i), E(m,\mathcal{T}_{i+1})$ are generated by the LDI. \cref{E2B2-div} is converted to the logarithmic domain and rewritten as the blurry-event loss $\mathcal{L}_{B\text{-}E}$,
\begin{equation}\label{log_be}
    \mathcal{L}_{B\text{-}E} = \|(\hat{B}_{i+1} - \hat{B}_{i}) - (\hat{E}(m,\mathcal{T}_{i+1}) - \hat{E}(m,\mathcal{T}_{i})) \|_1,
\end{equation}
where the top hat denotes logarithm, \eg, $\hat{B}_{i} = \operatorname{log}(B_i)$. 


\par 
\noindent \textbf{Blurry-Blurry Loss.}  
We revisit conventional blurring process \cref{blurry_latent} and reformulate as the discrete version, \ie,
\begin{equation}\label{reblur_evdi}
    \tilde{B}_i = \frac{1}{T} \int_{t\in \mathcal{T}_i} \hat{L}(t) dt \approx \frac{1}{M} \sum_{m=0}^{M-1} \hat{L}_{i} [m],
\end{equation}
where $\hat{L}_{i} [m]$ indicates the $m$-th latent image inside the exposure time of $B_i$ and $M$ denotes the total number of reconstructions. We formulate the blurry-blurry loss $\mathcal{L}_{B\text{-}B}$ between the re-blurred images $\tilde{B}_i, \tilde{B}_{i+1}$ and the original blurry inputs $B_{i}, B_{i+1}$ as
\begin{equation}\label{Blur-Blur}
    \mathcal{L}_{B\text{-}B} = \|\tilde{B}_i - B_i \|_1 + \|\tilde{B}_{i+1} - B_{i+1} \|_1.
\end{equation}

\par 
\noindent \textbf{Sharp-Event Loss.} 
Apart from the above constraints, the relation between sharp latent images and events can also be leveraged to supervise the reconstruction of consecutive latent frames. Based on \cref{latent_expEvent}, we have 
\begin{equation}\label{se_loss}
    \mathcal{N}(\Delta\hat{L}) = \mathcal{N}(J),
\end{equation}
where 
$\Delta\hat{L}\triangleq\hat{L}(t) - \hat{L}(m)$, $J\triangleq\int_f^t e(s) ds$ and $\mathcal{N}(\cdot)$ is the min/max normalization operator adopted in \cite{basics_paredes2021back}. The sharp-event loss $\mathcal{L}_{S\text{-}E}$ can be formulated as
\begin{equation}\label{se_loss_final}
    \mathcal{L}_{S\text{-}E} = \|\mathcal{M}(\mathcal{N}(\Delta \hat{L})) -  \mathcal{M}(\mathcal{N}(J)) \|_1,
\end{equation}
where $\mathcal{M}(\cdot)$ denotes a pixel-wise masking operator for $\mathcal{M}(\cdot) = 0$ only when there are no events.

\par
\noindent \textbf{Limitations.}
Two limitations primarily constrain the self-supervised losses in the EVDI~\cite{zhang2022unifying}: (i) The training efficiency of the EVDI is compromised owing to the reconstruction of multiple latent sharp frames $\hat{L}_{i} [m]$ in \cref{reblur_evdi} and $\hat{L}(t)$ in \cref{se_loss}; (ii) The performance of EVDI is hindered by the sharp-event loss, which assumes a globally consistent threshold within the exposure interval. While EVDI remains stable with synthetic event streams featuring globally uniform thresholds, it exhibits instability when applied to real-world event streams lacking such global consistency, as demonstrated in experiments presented in \cref{compare_with_SOTA}.        

\subsection{EVDI++}\label{EVDI++}
In addition to the LDI network, EVDI++ comprises a Learning-based Division Reconstruction (LDR) module, an Adaptive Parameter-free Fusion (APF) module, and enhanced self-supervised losses, as illustrated in \cref{framework}, which enable the generation of efficient and robust results across diverse scenes. Furthermore, we conduct a comprehensive comparison between EVDI++ and its previous version EVDI~\cite{zhang2022unifying} to highlight the advantages of our EVDI++.

\par 
\subsubsection{Reconstruction Network}
\par
\noindent \textbf{Exposure-Transferred Reconstruction.}
The latent sharp image $L(m)$ can be restored from the reference inputs $F\in\{B,I\}$ and the corresponding event integral $E(m,\mathcal{T})$ according to \cref{blurry_latent_event}. Considering the limitations of directly applying \cref{blurry_latent_event} for restoration (detailed in \cref{Subsec:Feasibility}), we propose a Learning-based Division Reconstruction (LDR) Network to reconstruct the refined sharp latent images $\bar{L}(m)$ by approximating \cref{blurry_latent_event},
\begin{equation}\label{LDR_L}
    \bar{L}(m) = \operatorname{LDR}(F,E(m,\mathcal{T})).
\end{equation}
Residual Dense Blocks (RDBs)~\cite{zhang2018residual} are employed to build our LDR as shown in \cref{fig:LDI_framework} (b).

Furthermore, since \cref{blurry_latent_event} can also be rewritten as $F=L(t)/E'(m,\mathcal{T})$ with $E'(\cdot)$ denotes the reciprocal of $E(\cdot)$, one can obtain
\begin{equation}\label{LDR_B}
    \bar{F} = \operatorname{LDR}(\bar{L}(m),E'(m,\mathcal{T})),
\end{equation}
where $\bar{F}$ indicates the reference image predicted from LDR. Thus, by employing the LDR module, our EVDI++ can efficiently restore the input reference image $\bar{F}$ to form cycle consistency, benefiting self-supervised learning and improving training efficiency. 
\par 
\noindent \textbf{Adaptive Parameter-free Fusion.} We can estimate two refined latent sharp images $\bar{L}_{i}(m)$ and $\bar{L}_{i+1}(m)$ at arbitrary timestamps $m$ by using LDR module as follows:
\begin{equation}\label{LDR_L_L}
\begin{aligned}
    \bar{L}_{i}(m) &= \operatorname{LDR}(F_{i},E(m,\mathcal{T}_{i})),
    \\
    \bar{L}_{i+1}(m) &= \operatorname{LDR}(F_{i+1},E(m,\mathcal{T}_{i+1})). 
\end{aligned}
\end{equation}
Moreover, a larger time interval between the target $m$ and reference $t_r$ timestamps indicates greater reconstruction challenges owing to accumulated errors, while a smaller interval suggests the opposite. The double integration $E(m,\mathcal{T}_i)$ calculated from the LDI implicitly captures information about motion complexity. To this end, we design an Adaptive Parameter-free Fusion (APF) module to obtain final latent sharp images $\tilde{L}(m)$ based on $\bar{L}_{i}(m)$, $\bar{L}_{i+1}(m)$  and corresponding $E(m,\mathcal{T}_i)$, $E(m,\mathcal{T}_{i+1})$, which can be formulated as
\begin{equation}
    \tilde{L}(m) =\left\{
	\begin{array}{ll}
	\bar{L}_{i}(m) & \text { if } m \in \mathcal{T}_i, 
    \vspace{.2em}
	\\
	W_{i} \cdot \bar{L}_{i}(m) + W_{i+1} \cdot \bar{L}_{i+1}(m) & \text { if } m \in \mathcal{T}_{i\rightarrow i+1},
    \vspace{.2em}
	\\
	\bar{L}_{i+1}(m) & \text { if } m \in \mathcal{T}_{i+1},
	\end{array}\right.
\end{equation}
where $W=\operatorname{Softmax}(|\operatorname{log}(E(m,\mathcal{T}_{i+1}))|,|\operatorname{log}(E(m,\mathcal{T}_{i}))|)$ are weights with $W=[W_i,W_{i+1}]$, and $|\cdot|$ means absolute value. Utilizing operations $\operatorname{log}$ and $|\cdot|$ is motivated by the fact that $E(m,\mathcal{T})=1$ signifies no luminance change, whereas $E(m,\mathcal{T})>1$ and $E(m,\mathcal{T})<1$ both indicate brightness changes, with greater deviations from 1 meaning more substantial variations. Furthermore, greater luminance differences often result in more pronounced cumulative errors, making smaller softmax values indicative of a more reliable estimation of $L(m)$ based on the corresponding events. Additionally, for $m\in \mathcal{T}_{i}$ and $m\in \mathcal{T}_{i+1}$, the final results $\tilde{L}_(m)$ directly align with $\bar{L}_{i}(m)$ and $\bar{L}_{i+1}(m)$, respectively, owing to the pronounced cumulative errors in $\bar{L}_{i+1}(m)$ and $\bar{L}_{i}(m)$ when $m\in \mathcal{T}_{i}$ and $m\in \mathcal{T}_{i+1}$. 

Our previous work, \ie, EVDI~\cite{zhang2022unifying}, employs a fusion network with learnable parameters and utilizes inputs from $L_i(m)$, $L_{i+1}(m)$, $L_{i+1}^{i}(m)$, $E(m,\mathcal{T}_{i})$ and $E(m,\mathcal{T}_{i+1})$, which proves beneficial only for reconstructing images outside of $\mathcal{T}_{i}$ and $\mathcal{T}_{i+1}$ while being detrimental for images within these intervals, as shown in the first row of \cref{EVDI_fusion}. In contrast, our APF module proposed in EVDI++ achieves excellent performance across the entire exposure intervals $t\in \mathcal{T}_{i+1}^{i}$ without any learnable parameters, as shown in the second row of \cref{EVDI_fusion}. 

\begin{table*}[htb]
\centering
\renewcommand{\arraystretch}{1.2}
\caption{Comparison between EVDI and EVDI++. Note that {training time is evaluated on the REDS dataset.}}
\begin{tabular}{c|cccc|cc|c|c|c}
\hline
\multirow{2}{*}{Method} & \multicolumn{4}{c|}{\begin{tabular}[c]{@{}c@{}}Self-Supervised loss\end{tabular}}  & \multicolumn{2}{c|}{Fusion Module}      & \multirow{2}{*}{Model Size} & \multirow{2}{*}{Training Time} & \multirow{2}{*}{FI} \\ \cline{2-7}
                        & \multicolumn{1}{c}{$\mathcal{L}_{B\text{-}E}$} & \multicolumn{1}{c}{$\mathcal{L}_{S\text{-}E}$} & \multicolumn{1}{c}{$\mathcal{L}_{B\text{-}B}$} & $\mathcal{L}_{S\text{-}B}$ & \multicolumn{1}{c}{\#Param.} & Effective Range &                     &                             &                                \\ \hline
EVDI                    & \multicolumn{1}{c}{\cref{log_be}}    & \multicolumn{1}{c}{\cref{se_loss_final}}    & \multicolumn{1}{c}{\cref{Blur-Blur}}      & \multicolumn{1}{c|}{/}           &  0.221 M   &     $t\in \mathcal{T}_{i\rightarrow{i+1}}$       &               0.396 M                        &              5760 s/epoch    &   \XSolidBrush                    \\ \hline
EVDI++                  & \multicolumn{1}{c}{\cref{be_LDR_L}}    & \multicolumn{1}{c}{/}    & \multicolumn{1}{c}{\cref{Blur_Blur_evdi++}}       & \multicolumn{1}{c|}{\cref{SB_loss}}      &   0 M  &       $t\in \mathcal{T}_{i}^{i+1}$                      &      0.357 M                       &      360 s/epoch       &  \Checkmark                         \\ \hline
\end{tabular}
\vspace{-1em}
\label{tab:compare_evdi_evdi++}
\end{table*}

\subsubsection{Self-supervised Learning}\label{selflearning}
The proposed self-supervised learning framework of our EVDI++ consists of three different losses including blurry-event $\mathcal{L}_{B\text{-}E}$, sharp-blurry $\mathcal{L}_{S\text{-}B}$, and blurry-blurry $\mathcal{L}_{B\text{-}B}$ constraints. Compared to its previous version, EVDI, the sharp-event loss is omitted from EVDI++ owing to its strong assumption regarding global threshold consistency, while a novel sharp-blurry loss is introduced to ensure brightness consistency. Furthermore, we optimize the reconstruction process for the blurry images $\bar{B}_{i}$ and $\bar{B}_{i+1}$, and the sharp latent images $\bar{L}_{i}$ and $\bar{L}_{i+1}$, enhancing training efficiency and stability through the incorporation of the LDR module proposed in EVDI++.

\par 
\noindent \textbf{Blurry-Event Loss.} 
The original blurry-event loss in \cref{log_be} of the EVDI is computed in the $\operatorname{log}$ domain, which differs from the other losses in linear domains, leading to disparate convergence rates during training. Consequently, the EVDI employs a two-stage training approach to alleviate this issue, which relies on manual tuning of loss weights and lacks flexibility. To this end, we construct the blurry-event loss in the linear domain by using the LDR module, which can be formulated as
\begin{equation}\label{be_LDR_L}
    \mathcal{L}_{B\text{-}E} = \|\operatorname{LDR}(B_i,E(m,\mathcal{T}_i))-\operatorname{LDR}(B_{i+1},E(m,\mathcal{T}_{i+1})) \|_1.
\end{equation}

\par 
\noindent \textbf{Sharp-Blurry Loss.} 
The brightness values in the static area of the refined sharp results $\bar{L}_{i}(m)$ and $\bar{L}_{i+1}(m)$ are consistent with the intensities of the input blurry images $B_{i}$ and $B_{i+1}$, respectively. Additionally, leveraging the intra-frame motion information encoded in events enables effective differentiation between static and dynamic scene areas. Thus, we introduce a sharp-blurry loss to ensure the brightness consistency between the blurry and sharp images by 
\begin{equation}\label{SB_loss}
\begin{aligned}
    \mathcal{L}_{S\text{-}B} &= \| (\bar{L}_{i}(m)-B_{i})*\mathcal{M}(\mathcal{E}_{\mathcal{T}_{i}})\|_{1}
    \\
    &+ \|(\bar{L}_{i+1}(m)-B_{i+1})*\mathcal{M}(\mathcal{E}_{\mathcal{T}_{i+1}}) \|_1,
\end{aligned}
\end{equation}
where $\mathcal{M}(\cdot)$ represents a mask operator, which assigns a value of 1 to static pixels (without emitted events) and a value of 0 to dynamic pixels (with emitted events). 
 
\par 
\noindent \textbf{Blurry-Blurry Loss.} 
According to \cref{reblur_evdi}, the training process of EVDI is notably time-consuming and inefficient, as it relies on the generation of a sequence of latent sharp images for the reconstruction of re-blurred blurry images $\tilde{B}_i$ and $\tilde{B}_{i+1}$. In contrast, EVDI++ employs an LDR scheme that directly reconstructs blurry images $\bar{B}_i$ and $\bar{B}_{i+1}$ to adapt to real-world scenarios and enhance training efficiency, which can be formulated as
\begin{equation}\label{LDR_reblur}
\begin{aligned}
    \bar{B}_{i} &= \operatorname{LDR}(\bar{L}_{i+1}(m),E'(m,\mathcal{T}_{i})),
    \\
    \bar{B}_{i+1} &= \operatorname{LDR}(\bar{L}_{i}(m),E'(m,\mathcal{T}_{i+1})). 
\end{aligned}
\end{equation}

We utilize $\bar{L}_{i+1}(m)$ and $\bar{L}_{i}(m)$ to reconstruction $\bar{B}_i$ and $\bar{B}_{i+1}$, respectively. This cross-supervision approach leverages the similarity between $\bar{L}_i(m)$ and $\bar{L}_{i+1}(m)$, mitigating the impact of identity mappings and fully harnessing the complementarities of adjacent reference blurry images $B_{i}$ and $B_{i+1}$ that embed varying degrees of sharpness information.   
 
The formulation of the blurry-blurry loss $\mathcal{L}_{B-B}$ in our proposed EVDI++ can be rewritten as
\begin{equation}\label{Blur_Blur_evdi++}
    \mathcal{L}_{B\text{-}B} = \|\bar{B}_i - B_i \|_1 + \|\bar{B}_{i+1} - B_{i+1} \|_1.
\end{equation}
\par 
\noindent \textbf{Total Loss.} Finally, the total self-supervised framework can be summarized as follows
\begin{equation}\label{total_loss}
  \mathcal{L} = \alpha \mathcal{L}_{B\text{-}E} + \beta \mathcal{L}_{S\text{-}B} + \gamma \mathcal{L}_{B\text{-}B},
\end{equation}
with $\alpha, \beta,\gamma$ denoting the balancing parameters.

\subsubsection{Comparative Analysis with EVDI}
As shown in \cref{tab:compare_evdi_evdi++}, we perform an extensive comparative analysis between the proposed EVDI++ and its prior work, \ie, EVDI~\cite{zhang2022unifying}, to highlight the advantages inherent in EVDI++. 

\noindent \textbf{Self-Supervised Losses.} Regarding the self-supervised losses in the EVDI++, we eliminate the sharp-event loss $\mathcal{L}_{S\text{-}E}$ in the EVDI, which relies on a strong assumption of global threshold consistency, and introduce sharp-blurry loss to enforce brightness consistency. Additionally, we improve the reconstruction of the re-blurred blurry images $\bar{B}$ in the blurry-blurry loss $\mathcal{L}_{B\text{-}B}$ by incorporating the LDR module.

\noindent \textbf{Fusion Method.} As shown in \cref{EVDI_fusion}, the proposed APF module achieves superior performance across all exposure intervals $t \in \mathcal{T}_{i+1}^{i}$ without requiring learnable parameters. This addresses the limitation in EVDI, where the parameterized fusion network is only effective in the middle of the exposure range $t \in \mathcal{T}_{i \rightarrow {i+1}}$.

\noindent \textbf{Task Capabilities.} EVDI++ exhibits superior generality compared to its previous version, \ie, EVDI~\cite{zhang2022unifying}. While EVDI is specialized for processing blurry inputs, EVDI++ can handle both blurry and sharp images, showcasing its enhanced versatility. 

\noindent \textbf{Model Size and Training Speed.} As shown in \cref{tab:compare_evdi_evdi++}, the proposed EVDI++ exhibits a 16-fold improvement in training speed (360 \vs 5760 s/epoch) and has fewer parameters (0.357 \vs 0.396 M).

\section{Experiments and Analysis}
In this section, the performance of our \myname\ is evaluated on Motion Deblurring (MD), Frame Interpolation (FI), and Blurry Video Enhancement (BVE). We first present the experimental settings in \cref{exp_set}, including details on datasets and implementations. Afterward, quantitative and qualitative comparisons of our \myname\ are made to the state-of-the-art methods on restorations of latent sharp images in \cref{compare_with_SOTA}. We explore the effectiveness of our network architecture and loss functions through ablations in \cref{exp_ablation}. The quantitative comparisons on the real-world ColorRBE dataset are described in \cref{exp_hrbe}. Finally, we elucidate the complexity of performance analysis for each task in \cref{sec_complexity}.

\subsection{Experimental Settings}\label{exp_set}
\subsubsection{Datasets} 

We evaluate the proposed method with three different datasets, including synthetic and real-world ones. 

\par 
\textbf{REDS:} We build a purely synthetic dataset based on the REDS dataset \cite{gopro_nah2019ntire}. We first downsample and crop the images to size $160\times320$ and then increase the frame rate by interpolating 7 images between consecutive frames using RIFE \cite{huang2022real}. Finally, we generate both blurry frames and events based on the high frame-rate sequences, where the blurry frames are obtained by averaging a specific number of sharp images, and events are simulated by ESIM \cite{esim_rebecq2018esim}.

\par 
\textbf{ColorDVS:} The ColorDVS dataset~\cite{kim2022event} captures real-world events and sharp frames in slow motion by using the DAVIS346c event camera. Moreover, we synthesize blurry frames using the same manner as the REDS dataset. The splitting of the training and testing sets for the real-world events dataset is identical to that of the official open-sourced ColorDVS dataset.

\par
\textbf{ColorRBE:} In contrast to the existing datasets with synthetic blurry images or events, we use a DAVIS346c camera to construct a Color Real-world Blurry images and Events (ColorRBE) dataset with diversity in terms of scenes, \ie, indoor and outdoor, and motion types, \ie, camera and object motion, which is used to verify adaptivity to the real-world scenario of our \myname.
\begin{table*}[th]
\centering
\small
\caption{Quantitative comparison of our proposed EVDI and EVDI++ with the state-of-the-art MD methods. Evaluations are conducted for single (middle) frame and sequence restoration, respectively. The best performance is in \textbf{bold} and the second best is \underline{underlined}. The symbol / denotes infeasible to reconstruct sharp sequences. The columns "Event" and "SSL" respectively denote the utilization of events and the self-supervised learning within the method.}
\vspace{-3mm}
\begin{tabular}{lccccccccccccc}
\hline
\multirow{3}{*}{Method} & \multicolumn{5}{c}{Single Frame Reconstruction}                                &  & \multicolumn{5}{c}{Seven Frames Reconstruction}                              & \multirow{3}{*}{Event} & \multirow{3}{*}{SSL} \\ \cline{2-6} \cline{8-12}
                        & \multicolumn{2}{c}{REDS-Sim} &  & \multicolumn{2}{c}{ColorDVS-Real}          &  & \multicolumn{2}{c}{REDS-Sim} &  & \multicolumn{2}{c}{ColorDVS-Real}         &                           \\ \cline{2-3} \cline{5-6} \cline{8-9} \cline{11-12}
                        & PSNR $\uparrow$       & SSIM $\uparrow$       &  & PSNR $\uparrow$          & SSIM $\uparrow$           &  & PSNR $\uparrow$       & SSIM $\uparrow$       &  & PSNR $\uparrow$          & SSIM $\uparrow$          &                           \\ \hline
LEVS~\cite{LEVS_jin2018learning}                    & 23.10       & 0.6542      &  & 25.43          & 0.7023          &  & 20.84       & 0.5473      &  & 25.35          & 0.7021     &  \XSolidBrush  &  \XSolidBrush                  \\
MotionETR~\cite{zhang2021exposure}                   & 23.67       & 0.6889      &  & 16.84          & 0.5838          &  & 20.43           & 0.5283           &  & 15.92              & 0.5189              & \XSolidBrush  &  \XSolidBrush                   \\
MPRNet~\cite{zamir2021multi}              & 23.24       & 0.6738      &  & 22.24          & 0.6831          &  & /       & /      &  & /          & /         & \XSolidBrush    &  \XSolidBrush                \\
DSTNet*~\cite{Pan_2023_CVPR}              & 23.27       & 0.6721      &  & 29.31          & 0.8399          &  & /      & /  &  & /  & /         & \XSolidBrush    &  \XSolidBrush                 \\
\hline
EDI~\cite{edi_pan2019bringing}                 & 21.90       & 0.6684      &  & 16.62          & 0.4266          &  & 21.29       & 0.6402      &  & 16.58          & 0.4219         & \Checkmark  &  \XSolidBrush                       \\
eSL-Net~\cite{esl_wang2020event}                & 16.76       & 0.5293      &  & 18.94          & 0.6647          &  & 17.80       & 0.5655      &  & 20.06          & 0.7488         & \Checkmark  &  \XSolidBrush                       \\
LEDVDI~\cite{ledvdi_lin2020learning} & 26.04       & 0.8681      &  & 25.06          & 0.7807          &  & 25.38       & 0.8567      &  & 25.07          & 0.7842         & \Checkmark  &  \XSolidBrush                       \\ 
RED-Net~\cite{red_xu2021motion} & 26.42       & 0.8788      &  & 20.39          & 0.6897          &  & 25.14       & 0.8587      &  & 20.76          & 0.6939         & \Checkmark  &  \Checkmark                       \\
RED-Net*~\cite{red_xu2021motion} & 30.83       & 0.8971      &  & 31.35          & 0.8862          &  & \underline{30.58}       & 0.8909      &  & \underline{31.17}          & \underline{0.8836}         & \Checkmark  &  \Checkmark                      \\
EFNet~\cite{sun2022event} & 26.49       & 0.8736      &  & 20.92          & 0.6464          &  & /       & /      &  & /          & /         & \Checkmark  &  \XSolidBrush                       \\
EFNet*~\cite{sun2022event} & \textbf{31.17}       & \textbf{0.9212}      &  & \textbf{34.86}          & \textbf{0.9339}          &  & /       & /      &  & /          & /         & \Checkmark  &  \XSolidBrush                       \\
\hline
EVDI (Ours)               & 30.85            & 0.9130          &  & 26.94 & 0.8117 &  & 30.40           & \underline{0.9058}        &  & 26.19 & 0.7965 & \Checkmark &  \Checkmark         \\ 
EVDI++ (Ours)               & \underline{31.11}            & \underline{0.9132}            &  & \underline{32.37} & \underline{0.8879} &  & \textbf{30.74}            & \textbf{0.9074}            &  & \textbf{32.05} & \textbf{0.8853} & \Checkmark & \Checkmark          \\ 
\hline
\end{tabular}
\label{tab:deblur}
\vspace{-3mm}
\end{table*}
\def\ssxxsone{(-0.42, -0.99)} 
\def\ssevdisone{(-0.45, -0.99)}
\def\ssyysone{(1.67, -0.43)} 

\def\ssxxstwo{(-1.1,.8)} 
\def\ssyystwo{(1.67,-0.49)} 

\def\ssizz{1.5cm} 
\def\sswidth{0.19\textwidth} 
\def\ssmag{3}
\def\scc{(-1.72,-1.15)}
\def\sccone{(-1.73,1.10)}

\def\ssizzone{1.5cm} 
\def\sswidthone{0.19\textwidth} 
\def\ssmagone{3}

\newcommand{\newimgfont}{\fontsize{7}{12}\selectfont}

\begin{figure*}[!thb]
	\centering

\begin{tikzpicture}[spy using outlines={green,magnification=\ssmagone,size=\ssizzone},inner sep=0]
		\node {\includegraphics[width=\sswidth]{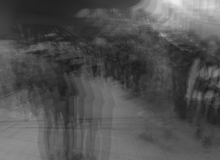}};
		\spy on \ssxxsone in node [left] at \ssyysone;
		\node [anchor=west, font=\newimgfont] at \sccone {\textcolor{yellow}{\bf Blurry Input}};
	\end{tikzpicture} 
    \begin{tikzpicture}[spy using outlines={green,magnification=\ssmagone,size=\ssizzone},inner sep=0]
		\node {\includegraphics[width=\sswidth,trim={0 0 0 0},clip,cframe=black .005mm]{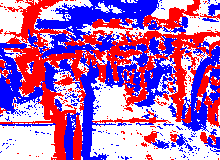}};
		\spy on \ssxxsone in node [left] at \ssyysone;
		\node [anchor=west, font=\newimgfont] at \sccone {\textcolor{black}{\bf Events}};
	\end{tikzpicture} 
	\begin{tikzpicture}[spy using outlines={green,magnification=\ssmagone,size=\ssizzone},inner sep=0]
		\node {\includegraphics[width=\sswidth]{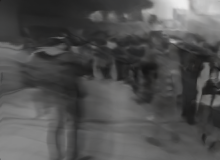}};
		\spy on \ssxxsone in node [left] at \ssyysone;
		\node [anchor=west, font=\newimgfont] at \sccone {\textcolor{yellow}{\bf MotionETR}};
	\end{tikzpicture} 
\begin{tikzpicture}[spy using outlines={green,magnification=\ssmagone,size=\ssizzone},inner sep=0]
		\node {\includegraphics[width=\sswidth]{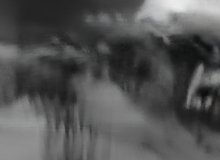}};
		\spy on \ssxxsone in node [left] at \ssyysone;
		\node [anchor=west, font=\newimgfont] at \sccone {\textcolor{yellow}{\bf MPRNET}};
	\end{tikzpicture} 
 \begin{tikzpicture}[spy using outlines={green,magnification=\ssmagone,size=\ssizzone},inner sep=0]
		\node {\includegraphics[width=\sswidth]{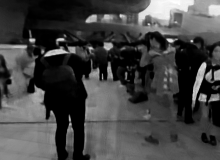}};
		\spy on \ssxxsone in node [left] at \ssyysone;
		\node [anchor=west, font=\newimgfont] at \sccone {\textcolor{yellow}{\bf LEDVDI}};
	\end{tikzpicture} 

    \vspace{.5mm}

    \begin{tikzpicture}[spy using outlines={green,magnification=\ssmagone,size=\ssizzone},inner sep=0]
		\node {\includegraphics[width=\sswidth]{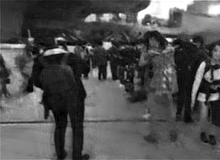}};
		\spy on \ssxxsone in node [left] at \ssyysone;
		\node [anchor=west, font=\newimgfont] at \sccone {\textcolor{yellow}{\bf RED-Net}};
	\end{tikzpicture} 
    \begin{tikzpicture}[spy using outlines={green,magnification=\ssmagone,size=\ssizzone},inner sep=0]
		\node {\includegraphics[width=\sswidth]{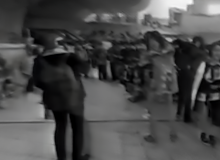}};
		\spy on \ssxxsone in node [left] at \ssyysone;
		\node [anchor=west, font=\newimgfont] at \sccone {\textcolor{yellow}{\bf EFNet}};
	\end{tikzpicture} 
	\begin{tikzpicture}[spy using outlines={green,magnification=\ssmagone,size=\ssizzone},inner sep=0]
		\node {\includegraphics[width=\sswidth]{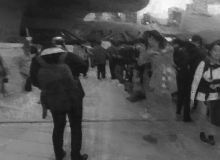}};
		\spy on \ssevdisone in node [left] at \ssyysone;
		\node [anchor=west, font=\newimgfont] at \sccone {\textcolor{yellow}{\bf EVDI}};
	\end{tikzpicture} 
	\begin{tikzpicture}[spy using outlines={green,magnification=\ssmagone,size=\ssizzone},inner sep=0]
		\node {\includegraphics[width=\sswidth]{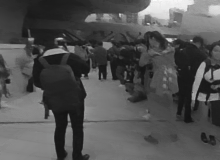}};
		\spy on \ssxxsone in node [left] at \ssyysone;
		\node [anchor=west, font=\newimgfont] at \sccone {\textcolor{yellow}{\bf EVDI++}};
	\end{tikzpicture} 
 \begin{tikzpicture}[spy using outlines={green,magnification=\ssmagone,size=\ssizzone},inner sep=0]
		\node {\includegraphics[width=\sswidth]{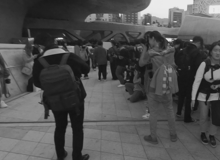}};
		\spy on \ssxxsone in node [left] at \ssyysone;
		\node [anchor=west, font=\newimgfont] at \sccone {\textcolor{yellow}{\bf GT}};
	\end{tikzpicture} 

\vspace{.5mm} 
    
	\begin{tikzpicture}[spy using outlines={green,magnification=\ssmag,size=\ssizz},inner sep=0]
		\node {\includegraphics[width=\sswidth]{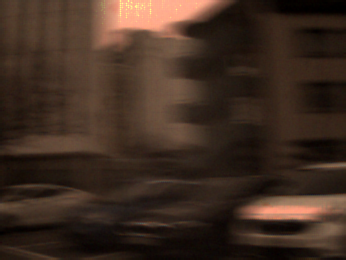}};
		\spy on \ssxxstwo in node [left] at \ssyystwo;
	\end{tikzpicture} 
    \begin{tikzpicture}[spy using outlines={green,magnification=\ssmag,size=\ssizz},inner sep=0]
		\node {\includegraphics[width=\sswidth,trim={0 0 0 0},clip,cframe=black .005mm]{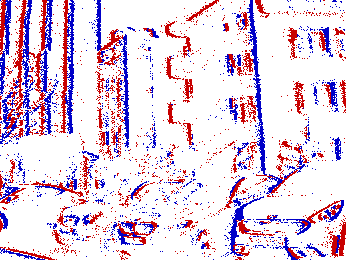}};
		\spy on \ssxxstwo in node [left] at \ssyystwo;
	\end{tikzpicture} 
	\begin{tikzpicture}[spy using outlines={green,magnification=\ssmag,size=\ssizz},inner sep=0]
		\node {\includegraphics[width=\sswidth]{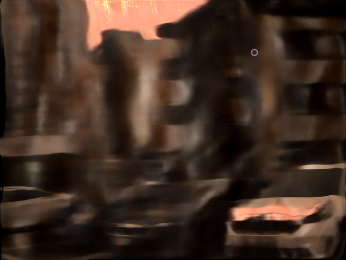}};
		\spy on \ssxxstwo in node [left] at \ssyystwo;
	\end{tikzpicture} 
\begin{tikzpicture}[spy using outlines={green,magnification=\ssmag,size=\ssizz},inner sep=0]
		\node {\includegraphics[width=\sswidth]{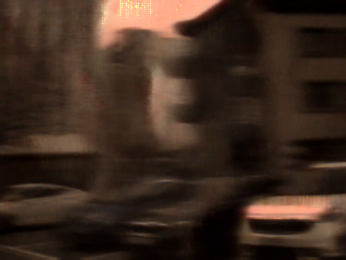}};
		\spy on \ssxxstwo in node [left] at \ssyystwo;
	\end{tikzpicture} 
 \begin{tikzpicture}[spy using outlines={green,magnification=\ssmag,size=\ssizz},inner sep=0]
		\node {\includegraphics[width=\sswidth]{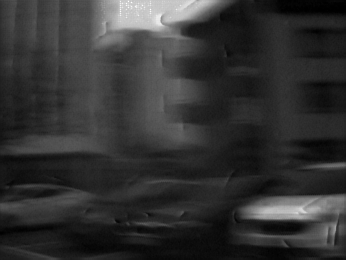}};
		\spy on \ssxxstwo in node [left] at \ssyystwo;
	\end{tikzpicture} 

    \vspace{.5mm}

    \begin{tikzpicture}[spy using outlines={green,magnification=\ssmag,size=\ssizz},inner sep=0]
		\node {\includegraphics[width=\sswidth]{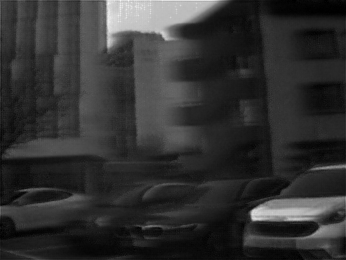}};
		\spy on \ssxxstwo in node [left] at \ssyystwo;
	\end{tikzpicture} 
    \begin{tikzpicture}[spy using outlines={green,magnification=\ssmag,size=\ssizz},inner sep=0]
		\node {\includegraphics[width=\sswidth]{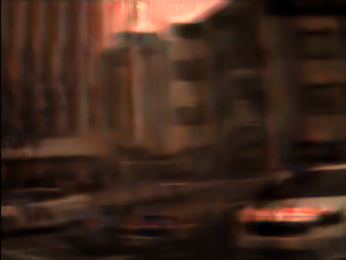}};
		\spy on \ssxxstwo in node [left] at \ssyystwo;
	\end{tikzpicture} 
	\begin{tikzpicture}[spy using outlines={green,magnification=\ssmag,size=\ssizz},inner sep=0]
		\node {\includegraphics[width=\sswidth]{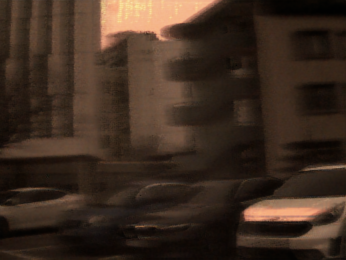}};
		\spy on \ssxxstwo in node [left] at \ssyystwo;
	\end{tikzpicture} 
	\begin{tikzpicture}[spy using outlines={green,magnification=\ssmag,size=\ssizz},inner sep=0]
		\node {\includegraphics[width=\sswidth]{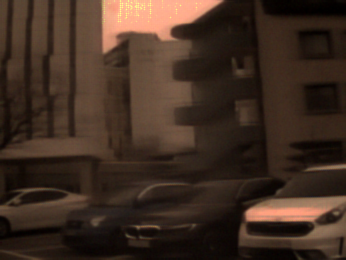}};
		\spy on \ssxxstwo in node [left] at \ssyystwo;
	\end{tikzpicture} 
 \begin{tikzpicture}[spy using outlines={green,magnification=\ssmag,size=\ssizz},inner sep=0]
		\node {\includegraphics[width=\sswidth]{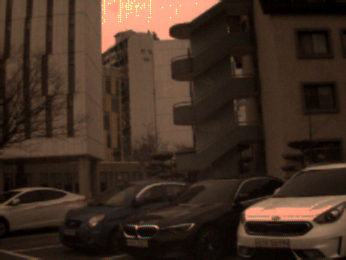}};
		\spy on \ssxxstwo in node [left] at \ssyystwo;
	\end{tikzpicture} 
 
	\caption{Qualitative comparisons of different deblurring methods on the REDS (top two rows) and ColorDVS (bottom two rows) datasets. Details are zoomed in for a better view. From left-top to right-bottom are, respectively, blurry images, events, and deblurring results by MotionETR~\cite{zhang2021exposure}, MPRNet~\cite{zamir2021multi}, LEDVID~\cite{ledvdi_lin2020learning}, RED-Net~\cite{red_xu2021motion}, EF-Net~\cite{sun2022event}, the proposed EVDI and EVDI++, and Ground Truth.}
	\label{exp_md}
\end{figure*}
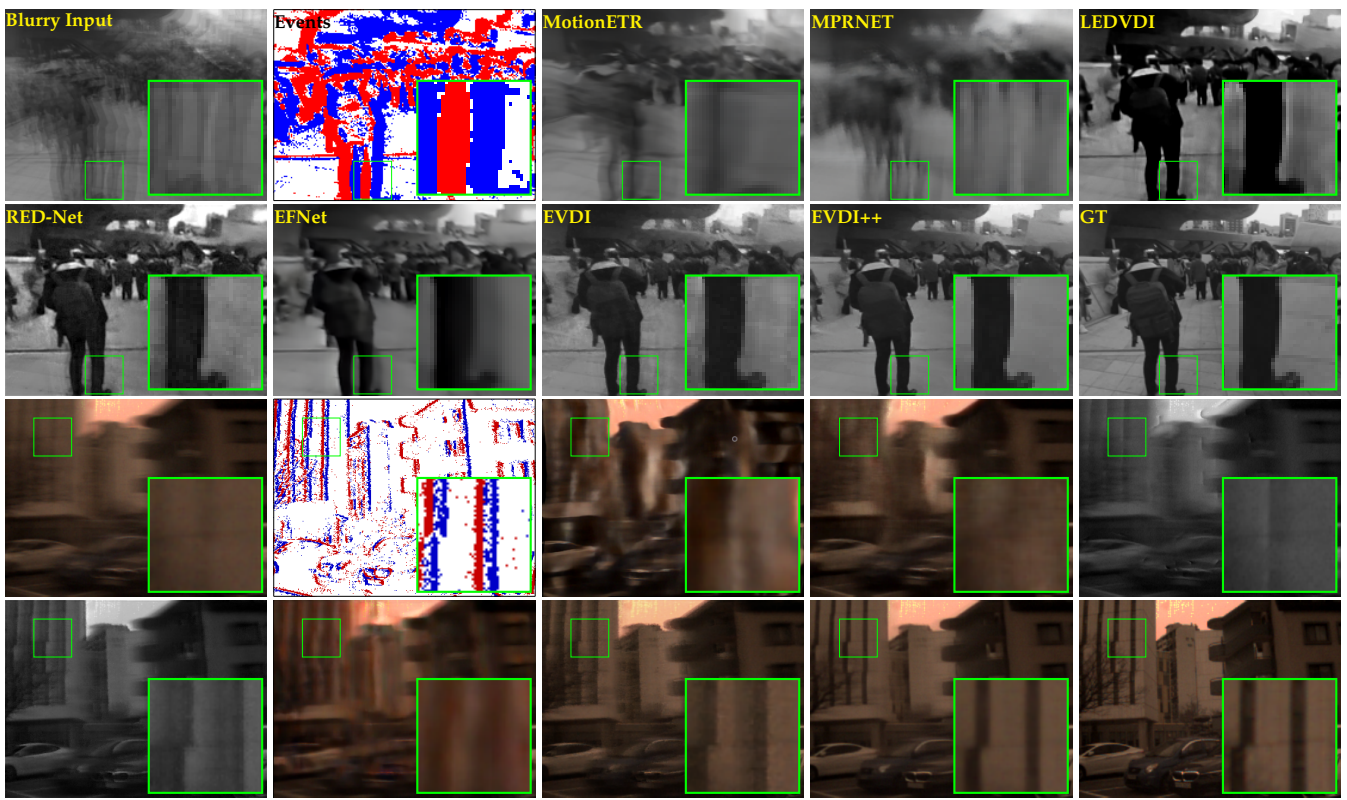
\subsubsection{Implementation Details and Evaluation Metrics}

\myname\ is implemented using PyTorch and trained on NVIDIA GeForce RTX 2080 Ti GPUs with a batch size of 4 by default. The Adam optimizer \cite{kingma2014adam} is employed, accompanied by the SGDR \cite{loshchilov2016sgdr} schedule, where the parameter $T_{max}$ is set to 200. We set the number of temporal bins $N=16$ for LDI inputs and randomly crop the images to $128\times 128$ patches for training. The weighting factors $\alpha$, $\beta$, and $\gamma$ in \cref{total_loss} are set to 1, 0.5, and 1, respectively. We train separate EVDI\text{++} models on the training sets of four datasets: REDS, ColorDVS, ColorRBE, and BS-ERGB, and then evaluate their performance on the corresponding testing sets. Additionally, we use Structural SIMilarity (SSIM)~\cite{wang2003multiscale} and Peak Signal-to-Noise Ratio (PSNR) as performance metrics for MD, FI, and BVE tasks.

\subsection{Comparison with the state-of-the-arts} \label{compare_with_SOTA}
\subsubsection{Performance of Motion Deblurring}
We first evaluate our \myname\ for the MD task. For deblurring, we synthesize one blurry image using 7 frames for the REDS dataset and 13 frames for the ColorDVS dataset. Performance assessment involves restoring 7 original frames per blurry image for both datasets. For the ColorDVS dataset, one image is skipped to align with the 7-frame restoration of the REDS dataset. All event-based methods were tested using the same number of events to ensure consistency. Our method is compared against state-of-the-art frame-based deblurring approaches, including LEVS \cite{LEVS_jin2018learning}, MotionETR~\cite{zhang2021exposure}, MPRNet~\cite{zamir2021multi}, DSTNet~\cite{Pan_2023_CVPR}, and various event-based methods, including EDI \cite{edi_pan2019bringing}, LEDVDI \cite{ledvdi_lin2020learning}, eSL-Net \cite{esl_wang2020event}, RED-Net \cite{red_xu2021motion}, and EFNet~\cite{sun2022event}.

\begin{table*}[th]
\centering
\small
\caption{Quantitative comparison of our proposed EVDI and EVDI++ with the state-of-the-art VFI methods. Evaluations are conducted for single frame and sequence interpolation, respectively. The best performance is in \textbf{bold} and the second best is \underline{underlined}.}
\vspace{-3mm}
\begin{tabular}{lccccccccccccc}
\hline
\multirow{3}{*}{Method} & \multicolumn{5}{c}{Single Frame Interpolation}                                &  & \multicolumn{5}{c}{Five Frames Interpolation}                              & \multirow{3}{*}{Event} & \multirow{3}{*}{SSL} \\ \cline{2-6} \cline{8-12}
                        & \multicolumn{2}{c}{REDS-Sim} &  & \multicolumn{2}{c}{ColorDVS-Real}          &  & \multicolumn{2}{c}{REDS-Sim} &  & \multicolumn{2}{c}{ColorDVS-Real}         &                           \\ \cline{2-3} \cline{5-6} \cline{8-9} \cline{11-12}
                        & PSNR $\uparrow$       & SSIM $\uparrow$       &  & PSNR $\uparrow$          & SSIM $\uparrow$           &  & PSNR $\uparrow$       & SSIM $\uparrow$       &  & PSNR $\uparrow$          & SSIM $\uparrow$          &                           \\ \hline
RIFE~\cite{huang2022real}                    & 20.19       & 0.5049      &  & 23.24          & 0.6445          &  & 21.07       & 0.5553      &  & 24.64          & 0.6837     &  \XSolidBrush  &  \XSolidBrush                  \\
EMA~\cite{zhang2023extracting}                    & 20.76       & 0.5325      &  & 24.58          & 0.7089          &  & 21.57           & 0.5805           &  & 24.29              & 0.6918              & \XSolidBrush  &  \XSolidBrush                   \\
UnSuperS~\cite{reda2019unsupervised}                   & 18.05       & 0.4427      &  & 21.62          & 0.6186          &  & 18.07       & 0.4420      &  & 21.96          & 0.6293         & \XSolidBrush    &  \Checkmark                \\
\hline
EDI~\cite{edi_pan2019bringing}                 & 16.60       & 0.3731      &  & 21.69          & 0.7274          &  & 18.02       & 0.4593      &  & 21.92          & 0.7384         & \Checkmark  &  \XSolidBrush                       \\
TimeLens~\cite{tulyakov2021time}                & \underline{25.24}       & \underline{0.8084}      &  & 27.92          &     0.8001      &  & \underline{25.76}       & \underline{0.8176}      &  & 27.86          & 0.7949         & \Checkmark  &  \XSolidBrush                       \\
REFID~\cite{sun2023event}                & 24.39       & 0.7812      &  & \underline{28.54}           &     \underline{0.8401}      &  & 24.34       & 0.7808      &  & \underline{28.51}          & \underline{0.8383}         & \Checkmark  &  \XSolidBrush                       \\
\hline
EVDI++ (Ours)               & \textbf{29.94}            & \textbf{0.8904}            &  & \textbf{31.74} & \textbf{0.8859} &  & \textbf{30.07}            & \textbf{0.8934}            &  & \textbf{32.83} & \textbf{0.8883} & \Checkmark & \Checkmark          \\ 
\hline
\end{tabular}
\label{tab:interpolation}
\vspace{-3mm}
\end{table*}
\input{Figs/Fig-interp}

\begin{table}[hbpt]
    \centering
    \caption{Quantitative comparison on BS-ERGB dataset of our method with frame- and hybrid frame+event-based methods. The ``\#Param.'' reports the number of parameters for each method.}
    \resizebox{\linewidth}{!}{
    \begin{tabular}{c|cccccccccc}
\hline
        \multirow{2}{*}{Method} & \multirow{2}{*}{Events} & \multirow{2}{*}{SSL} & \multirow{2}{*}{\#Param.} & \multicolumn{2}{c}{PSNR$\uparrow$} \\ \cline{5-6}
                &  &  &  & 1-SKIP & 3-SKIP \\ \hline
            FLAVR & \XSolidBrush & \XSolidBrush & 32.0M & 25.95    &   20.90      \\ 
             DAIN  & \XSolidBrush & \XSolidBrush & 24.3M & 25.20 &  21.40    \\
             SuperSlomo  & \XSolidBrush & \XSolidBrush & 28.5M & -/- &    22.48  \\
          QVI & \XSolidBrush & \XSolidBrush & - & -/-      &  23.20      \\
           TimeLens  & \Checkmark & \XSolidBrush & 72.2M &  28.36 & 27.58  \\
           TimeLens++  & \Checkmark & \XSolidBrush & 53.9M &   28.56   &  27.63 \\
           EVDI++ & \Checkmark & \Checkmark & 0.35M &    26.68   &   25.43 \\
           EVDI++ (Large) & \Checkmark & \Checkmark & 1.10M & 27.81 & 26.53  \\
           \hline
        
    \end{tabular}
    }
    \label{tab:interpolation_bsergb}
\end{table}

\noindent \textbf{Quantitative results.} The quantitative results of the single frame and sequence reconstruction on the REDS and ColorDVS datasets are shown in \cref{tab:deblur}. Both EVDI and \myname\ outperform the state-of-the-art MD methods by a large margin. Specifically, EVDI++ achieves an average improvement of 7.6/8.9 dB and 0.245/0.241 for PSNR and SSIM, respectively, in single-frame and sequence reconstruction on the REDS dataset, and 11.60/11.43 dB and 0.241/0.239 on the ColorDVS dataset. The event-based approaches, \ie, RED-Net~\cite{red_xu2021motion}, LEDVDI~\cite{ledvdi_lin2020learning}, and EF-Net~\cite{sun2022event} can obtain better performance than the frame-based methods, \ie, MotionETR~\cite{zhang2021exposure} and MPRNet~\cite{zamir2021multi}, owing to the introduction of the events. Despite the strong performance of existing methods, their quantitative results for both single-frame and sequence reconstruction on the REDS and ColorDVS datasets are notably lower than those achieved by our EVDI++. This is because our EVDI++ directly fits the target data distributions with the proposed self-supervised training approach, while other pre-trained methods struggle to generalize due to the distribution gap between the training and testing data. To further show the superiority of our EVDI++, we retrain the latest state-of-the-art models (DSTNet*, RED-Net*, and EF-Net*) on the REDS and ColorDVS training sets using their official code and default parameters. This retraining yielded significant improvements in PSNR and SSIM, particularly on the ColorDVS dataset. Further improvements could likely be achieved through model enhancements, such as increasing the number of parameters or employing specialized architectures. However, our self-supervised EVDI++ not only achieves outstanding performance, but also has the unique capability to reconstruct sharp video sequences at arbitrary timestamps throughout the entire exposure period, $T^{i}_{i+1}$, whereas other methods are restricted to reconstructing single or fixed-length sequences within discrete intervals, $T_{i}$ or $T_{i+1}$. Furthermore, \myname\ outperforms its previous version, EVDI, on both datasets with synthetic and real-world events, demonstrating the effectiveness of the exposure-transferred reconstruction and adaptive parameter-free fusion modules for motion deblurring. 

\def\ssxxsone{(0.5, -0.12)} 
\def\ssevdisone{(0.5, -0.12)}
\def\ssyysone{(-0.21, -0.47)} 

\def\ssxxstwo{(1.0,1.0)} 
\def\ssyystwo{(-0.19,-0.50)} 

\def\ssizz{1.5cm} 
\def\sswidth{0.195\textwidth} 
\def\ssmag{3}
\def\scc{(1.70,-1.15)}
\def\sccone{(-1.73,1.10)}

\def\ssizzone{1.5cm} 
\def\sswidthone{0.195\textwidth} 
\def\ssmagone{3}

\begin{figure*}[!thb]
	\centering

\begin{tikzpicture}[spy using outlines={green,magnification=\ssmagone,size=\ssizzone},inner sep=0]
		\node {\includegraphics[width=\sswidth]{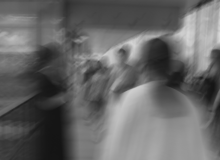}};
		\spy on \ssxxsone in node [left] at \ssyysone;
		\node [anchor=west, font=\newimgfont] at \sccone {\textcolor{yellow}{\bf Start Frame}};
	\end{tikzpicture} 
    \begin{tikzpicture}[spy using outlines={green,magnification=\ssmagone,size=\ssizzone},inner sep=0]
		\node {\includegraphics[width=\sswidth,trim={0 0 0 0},clip,cframe=black .005mm]{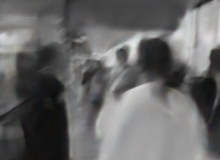}};
		\spy on \ssxxsone in node [left] at \ssyysone;
		\node [anchor=west, font=\newimgfont] at \sccone {\textcolor{yellow}{\bf Jin}};
	\end{tikzpicture} 
	\begin{tikzpicture}[spy using outlines={green,magnification=\ssmagone,size=\ssizzone},inner sep=0]
		\node {\includegraphics[width=\sswidth]{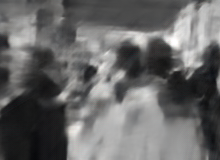}};
		\spy on \ssxxsone in node [left] at \ssyysone;
		\node [anchor=west, font=\newimgfont] at \sccone {\textcolor{yellow}{\bf Bin}};
	\end{tikzpicture} 
\begin{tikzpicture}[spy using outlines={green,magnification=\ssmagone,size=\ssizzone},inner sep=0]
		\node {\includegraphics[width=\sswidth]{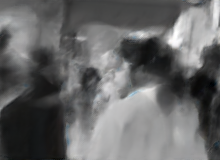}};
		\spy on \ssxxsone in node [left] at \ssyysone;
		\node [anchor=west, font=\newimgfont] at \sccone {\textcolor{yellow}{\bf DeMFI}};
	\end{tikzpicture} 
 \begin{tikzpicture}[spy using outlines={green,magnification=\ssmagone,size=\ssizzone},inner sep=0]
		\node {\includegraphics[width=\sswidth]{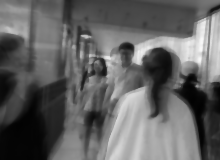}};
		\spy on \ssxxsone in node [left] at \ssyysone;
		\node [anchor=west, font=\newimgfont] at \sccone {\textcolor{yellow}{\bf EDI}};
	\end{tikzpicture} 

    \vspace{.5mm}

    \begin{tikzpicture}[spy using outlines={green,magnification=\ssmagone,size=\ssizzone},inner sep=0]
		\node {\includegraphics[width=\sswidth]{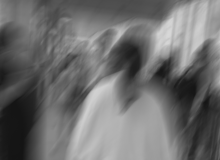}};
		\spy on \ssxxsone in node [left] at \ssyysone;
		\node [anchor=west, font=\newimgfont] at \sccone {\textcolor{yellow}{\bf End Frame}};
	\end{tikzpicture} 
    \begin{tikzpicture}[spy using outlines={green,magnification=\ssmagone,size=\ssizzone},inner sep=0]
		\node {\includegraphics[width=\sswidth]{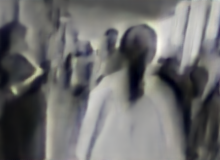}};
		\spy on \ssxxsone in node [left] at \ssyysone;
		\node [anchor=west, font=\newimgfont] at \sccone {\textcolor{yellow}{\bf REFID}};
	\end{tikzpicture} 
	\begin{tikzpicture}[spy using outlines={green,magnification=\ssmagone,size=\ssizzone},inner sep=0]
		\node {\includegraphics[width=\sswidth]{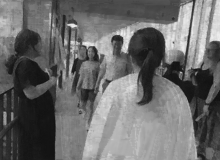}};
		\spy on \ssevdisone in node [left] at \ssyysone;
		\node [anchor=west, font=\newimgfont] at \sccone {\textcolor{yellow}{\bf EVDI}};
	\end{tikzpicture} 
	\begin{tikzpicture}[spy using outlines={green,magnification=\ssmagone,size=\ssizzone},inner sep=0]
		\node {\includegraphics[width=\sswidth]{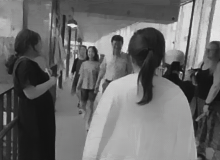}};
		\spy on \ssxxsone in node [left] at \ssyysone;
		\node [anchor=west, font=\newimgfont] at \sccone {\textcolor{yellow}{\bf EVDI++}};
	\end{tikzpicture} 
 \begin{tikzpicture}[spy using outlines={green,magnification=\ssmagone,size=\ssizzone},inner sep=0]
		\node {\includegraphics[width=\sswidth]{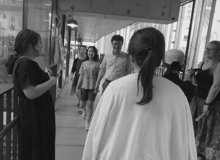}};
		\spy on \ssxxsone in node [left] at \ssyysone;
		\node [anchor=west, font=\newimgfont] at \sccone {\textcolor{yellow}{\bf GT}};
	\end{tikzpicture} 

\vspace{.5mm} 
    
	\begin{tikzpicture}[spy using outlines={green,magnification=\ssmag,size=\ssizz},inner sep=0]
		\node {\includegraphics[width=\sswidth]{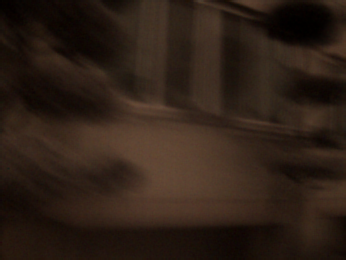}};
		\spy on \ssxxstwo in node [left] at \ssyystwo;
		\node [anchor=east, font=\newimgfont] at \scc {\textcolor{yellow}{\bf Start Frame}};
	\end{tikzpicture} 
    \begin{tikzpicture}[spy using outlines={green,magnification=\ssmag,size=\ssizz},inner sep=0]
		\node {\includegraphics[width=\sswidth,trim={0 0 0 0},clip,cframe=black .005mm]{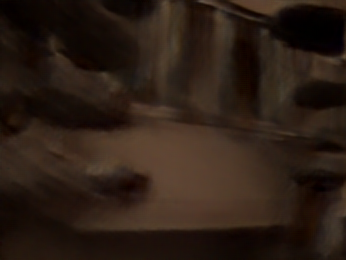}};
		\spy on \ssxxstwo in node [left] at \ssyystwo;
		\node [anchor=east, font=\newimgfont] at \scc {\textcolor{yellow}{\bf Jin}};
	\end{tikzpicture} 
	\begin{tikzpicture}[spy using outlines={green,magnification=\ssmag,size=\ssizz},inner sep=0]
		\node {\includegraphics[width=\sswidth]{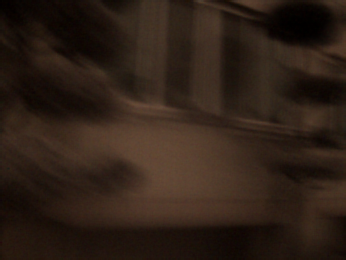}};
		\spy on \ssxxstwo in node [left] at \ssyystwo;
		\node [anchor=east, font=\newimgfont] at \scc {\textcolor{yellow}{\bf Bin}};
	\end{tikzpicture} 
\begin{tikzpicture}[spy using outlines={green,magnification=\ssmag,size=\ssizz},inner sep=0]
		\node {\includegraphics[width=\sswidth]{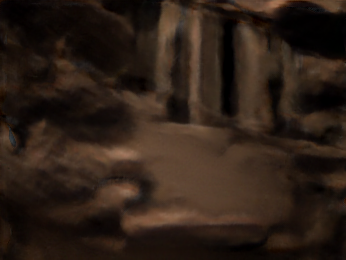}};
		\spy on \ssxxstwo in node [left] at \ssyystwo;
		\node [anchor=east, font=\newimgfont] at \scc {\textcolor{yellow}{\bf DeMFI}};
	\end{tikzpicture} 
 \begin{tikzpicture}[spy using outlines={green,magnification=\ssmag,size=\ssizz},inner sep=0]
		\node {\includegraphics[width=\sswidth]{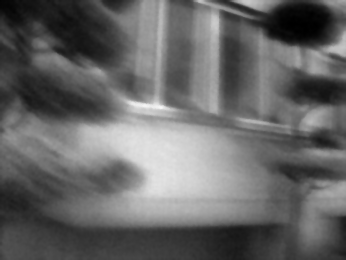}};
		\spy on \ssxxstwo in node [left] at \ssyystwo;
		\node [anchor=east, font=\newimgfont] at \scc {\textcolor{yellow}{\bf EDI}};
	\end{tikzpicture} 

    \vspace{.5mm}

    \begin{tikzpicture}[spy using outlines={green,magnification=\ssmag,size=\ssizz},inner sep=0]
		\node {\includegraphics[width=\sswidth]{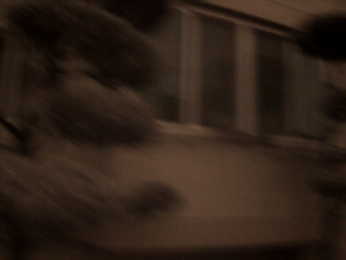}};
		\spy on \ssxxstwo in node [left] at \ssyystwo;
		\node [anchor=east, font=\newimgfont] at \scc {\textcolor{yellow}{\bf End Frame}};
	\end{tikzpicture} 
    \begin{tikzpicture}[spy using outlines={green,magnification=\ssmag,size=\ssizz},inner sep=0]
		\node {\includegraphics[width=\sswidth]{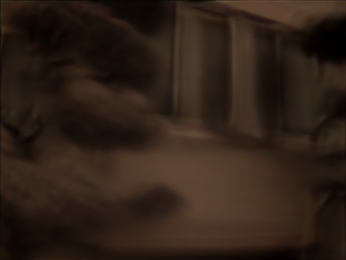}};
		\spy on \ssxxstwo in node [left] at \ssyystwo;
		\node [anchor=east, font=\newimgfont] at \scc {\textcolor{yellow}{\bf REFID}};
	\end{tikzpicture} 
	\begin{tikzpicture}[spy using outlines={green,magnification=\ssmag,size=\ssizz},inner sep=0]
		\node {\includegraphics[width=\sswidth]{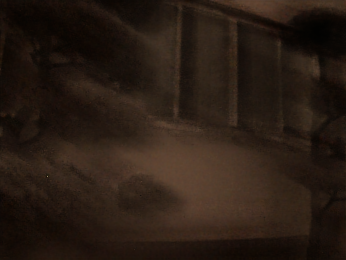}};
		\spy on \ssxxstwo in node [left] at \ssyystwo;
		\node [anchor=east, font=\newimgfont] at \scc {\textcolor{yellow}{\bf EVDI}};
	\end{tikzpicture} 
	\begin{tikzpicture}[spy using outlines={green,magnification=\ssmag,size=\ssizz},inner sep=0]
		\node {\includegraphics[width=\sswidth]{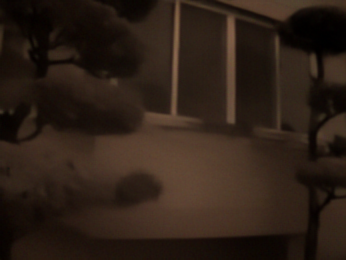}};
		\spy on \ssxxstwo in node [left] at \ssyystwo;
		\node [anchor=east, font=\newimgfont] at \scc {\textcolor{yellow}{\bf EVDI++}};
	\end{tikzpicture} 
 \begin{tikzpicture}[spy using outlines={green,magnification=\ssmag,size=\ssizz},inner sep=0]
		\node {\includegraphics[width=\sswidth]{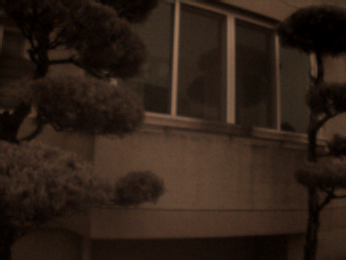}};
		\spy on \ssxxstwo in node [left] at \ssyystwo;
		\node [anchor=east, font=\newimgfont] at \scc {\textcolor{yellow}{\bf GT}};
	\end{tikzpicture} 
 
	\caption{Qualitative comparisons of different blurry video enhancement algorithms on the REDS (top two rows) and ColorDVS (bottom two rows) datasets.}
	\label{fig:exp_blurry_interp}
\end{figure*}

\noindent \textbf{Qualitative results.}
We further evaluate the performance qualitatively on the ColorDVS and REDS datasets, as shown in \cref{exp_md}. On the one hand, the results in the REDS dataset are shown in the top two rows of \cref{exp_md}. The results estimated by frame-based methods, \ie, MotionETR~\cite{zhang2021exposure} and MPRNet~\cite{zamir2021multi}, suffer from artifacts and distortions, degrading the overall quality of the deblurred results. The event-based methods, \ie, LEDVDI~\cite{ledvdi_lin2020learning}, RED-Net~\cite{red_xu2021motion}, and EFNet~\cite{sun2022event} can alleviate the artifacts by leveraging intra-frame motion information embedded in events, but the reconstructed details still suffer distortions. On the contrary, our EVDI and EVDI++ can obtain accurate reconstructions that exhibit the most similar appearances to the ground-truth sharp images. On the other hand, the bottom two rows of \cref{exp_md} present qualitative results on the ColorDVS dataset with {\it real-world} events. Frame-based methods are ineffective in recovering targets in a detailed way, \eg, {\it pinstripe}. Event-based methods, \ie, RED-Net and EFNet, can restore sharper contours of targets but suffer suboptimal solutions. Both EVDI and EVDI++ predict results with sharper edges and smoother surfaces than their competitors, demonstrating the effectiveness of our unified and self-supervised architecture. And EVDI++ still outperforms EVDI, producing fewer noises and artifacts, \eg, halo effects. Furthermore, compared to existing approaches that can only recover a single or limited number of latent sharp images inside $\mathcal{T}_i$, our proposed EVDI++ can restore latent sharp images of arbitrary timestamps both inside $\mathcal{T}_i$ (or $\mathcal{T}_{i+1}$) and $\mathcal{T}_{i\rightarrow {i+1}}$. The video reconstruction demonstration is available on our project page: \url{https://bestrivenzc.github.io/EVDI-plus-plus/}.

\subsubsection{Performance of Frame Interpolation}
We further evaluate our EVDI++ for the Frame Interpolation (FI) task. Unlike the 5-frame skips for sequences in the REDS dataset, we set 11-frame skips for the ColorDVS dataset due to its slow-motion characteristics. Performance assessment involves restoring 5 original frames between two sharp images for both datasets. For the ColorDVS dataset, one frame is skipped to align with the 5-frame restoration of the REDS dataset. Different FI methods are compared including the frame-based methods, \ie, EMA~\cite{zhang2023extracting}, RIFE~\cite{huang2022real}, and UnSuperSlomo~\cite{reda2019unsupervised}, and the event-based methods, \ie, EDI~\cite{edi_pan2019bringing} and TimeLens~\cite{tulyakov2021time}. Quantitative and qualitative results are respectively given in \cref{tab:interpolation} and \cref{fig:exp_com_interp}.

Regarding the quantitative results, we evaluate the performance of both single (middle frame) and sequence (all frames) interpolation on the REDS and ColorDVS datasets. Our proposed \myname\ significantly outperforms the state-of-the-art. 
Specifically, \myname\ achieves an average gain of 9.77/9.17 dB and 0.193/0.188 for PSNR and SSIM for single-frame and sequence reconstruction on the REDS dataset, respectively. The superiority is even more pronounced on the
ColorDVS dataset, where EVDI++ achieves 11.60/11.43
dB and 0.241/0.239 improvements in PSNR and SSIM
for frame and sequence reconstructions, respectively. 

For qualitative comparisons of the sequence interpolation, we show the sequence results on the ColorDVS dataset by five methods, \ie, EMA, RIFE, Timelens, EVDI, and \myname, as shown in \cref{fig:exp_com_interp}. EMA and RIFE are both frame-based FI methods and struggle with fast and complex motions, yielding blurred and distorted artifacts across ColorDVS datasets. Although the event-based method, \ie, TimeLens, improves precision on the ColorDVS dataset by incorporating event information, it remains vulnerable to distortions under large displacements. On the contrary, the proposed \myname\ interpolates frames with sharper details and smoother inter-frame transitions than its competitors on both datasets, demonstrating the effectiveness of integrating the FI task. \re{Additionally, the quantitative comparisons on BS-ERGB dataset~\cite{tulyakov2022time} with higher resolution are shown in \cref{tab:interpolation_bsergb}.}


Moreover, unlike EVDI, which is designed for blurry inputs, EVDI++ is capable of handling both blurry and sharp images, exhibiting superior generalization performance.  

\begin{table*}[th]
\centering
\small
\caption{Quantitative comparison of our proposed EVDI and EVDI++ with the state-of-the-art BVE methods. Evaluations are conducted for single frame and sequence interpolation, respectively. The symbol / and * denotes that it is infeasible to reconstruct sharp sequences and the retraining. The best performance is in \textbf{bold}, and the second best is \underline{underlined}.}
\vspace{-3mm}
\begin{tabular}{lccccccccccccc}
\hline
\multirow{3}{*}{Method} & \multicolumn{5}{c}{Single Frame Interpolation}                                &  & \multicolumn{5}{c}{Three Frames Interpolation}                              & \multirow{3}{*}{Event} & \multirow{3}{*}{SSL} \\ \cline{2-6} \cline{8-12}
                        & \multicolumn{2}{c}{REDS-Sim} &  & \multicolumn{2}{c}{ColorDVS-Real}          &  & \multicolumn{2}{c}{REDS-Sim} &  & \multicolumn{2}{c}{ColorDVS-Real}         &                           \\ \cline{2-3} \cline{5-6} \cline{8-9} \cline{11-12}
                        & PSNR $\uparrow$       & SSIM $\uparrow$       &  & PSNR $\uparrow$          & SSIM $\uparrow$           &  & PSNR $\uparrow$       & SSIM $\uparrow$       &  & PSNR $\uparrow$          & SSIM $\uparrow$          &                           \\ \hline
Jin~\cite{jin2019learning}                    & 19.16       & 0.4529      &  & 24.19          & 0.7110          &  & 19.50       & 0.4730      &  & 24.46          & 0.7144     &  \XSolidBrush  &  \XSolidBrush                  \\
Bin~\cite{shen2020blurry}                    & 18.48       & 0.4075      &  & 19.41          & 0.6122          &  & /           & /           &  & /              & /              & \XSolidBrush  &  \XSolidBrush                   \\
DeMFI~\cite{oh2022demfi}                   & 18.33       & 0.4128      &  & 22.65          & 0.6423          &  & 18.45       & 0.4324      &  & 22.74          & 0.6501         & \XSolidBrush    &  \Checkmark                \\
\re{DeMFI*}~\cite{oh2022demfi}                   & \re{24.86}       & \re{0.7256}      &  & \re{26.11}          & \re{0.7143}          &  & \re{24.81}       & \re{0.7248}      &  & \re{25.98}          & \re{0.7135}   & \re{\XSolidBrush}    &  \re{\Checkmark}       \\
\hline
EDI~\cite{edi_pan2019bringing}                 & 17.94       & 0.4625      &  & 22.12          & 0.7246          &  & 18.49       & 0.4862      &  & 22.04          & 0.7253         & \Checkmark  &  \XSolidBrush                       \\
REFID~\cite{sun2023event}                & 22.12      & 0.6418      &  & 18.55           &     0.5880      &  & 22.23       & 0.6375      &  & 18.66          & 0.5901         & \Checkmark  &  \XSolidBrush                       \\
\re{REFID*}~\cite{sun2023event}               & \re{27.06}      & \re{0.7963}      &  & \re{\underline{28.67}}           &     \re{\underline{0.7723}}     &  & \re{26.95}       & \re{0.7958}     &  & \re{\underline{28.58}}         & \re{\underline{0.7699}}    & \re{\Checkmark}  &  \re{\XSolidBrush}         \\
\hline
EVDI* (Ours)               & \underline{28.64}            & \underline{0.8701}          &  & 24.51 & 0.7577 &  & \underline{28.77}           & \underline{0.8731}        &  & 24.70 & 0.7628 & \Checkmark &  \Checkmark         \\ 
EVDI++ (Ours)                & \textbf{29.41}            & \textbf{0.8806}            &  & \textbf{31.60} & \textbf{0.8668} &  & \textbf{29.48}            & \textbf{0.8822}            &  & \textbf{31.64} & \textbf{0.8678} & \Checkmark & \Checkmark          \\ 
\hline
\end{tabular}
\label{tab:blur_interp}
\vspace{-3mm}
\end{table*}

\def\imwidth{0.192}

\def\sxxxl{(0.10,1.12)}
\def\sxxxrl{(1.28,-0.88)}

\def\sxxxxl{(0.10,1.12)}
\def\sxxxxrl{(1.28,-0.88)}

\def\ssyys{(1.75,-2.20)}
\def\ssyysr{(-0.03,-2.20)}

\def\ssyy{(-0.8,-0.85)}
\def\ssizz{1.7cm}
\def\sswidth{0.245\textwidth}
\def\ssmag{5}
\def\scc{(2.12,1.4)}
\begin{figure*}[!ht]
\footnotesize
	\centering
    	\begin{minipage}[t]{\imwidth\linewidth}
    		\centering
			\begin{tikzpicture}[spy using outlines={green,magnification=\ssmag,size=\ssizz},inner sep=0]
				\node {\includegraphics[width=\linewidth]{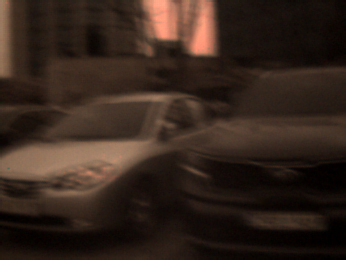}};
				\spy on \sxxxxl in node [left] at \ssyysr;
				\spy[red] on \sxxxxrl in node [left,red] at \ssyys;
				\end{tikzpicture}
			(a) Blurry Image \\
            (PSNR, SSIM)
            \vspace{0.5em}
                
			\begin{tikzpicture}[spy using outlines={green,magnification=\ssmag,size=\ssizz},inner sep=0]
				\node {\includegraphics[width=\linewidth]{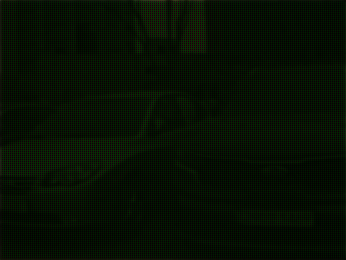}};
				\spy on \sxxxl in node [left] at \ssyysr;
				\spy[red] on \sxxxrl in node [left,red] at \ssyys;
				\end{tikzpicture}
			(f) w/o $\mathcal{L}_{S\text{-}B}$ \\
            (14.83, 0.2485) \vspace{0.3em}
    	\end{minipage}%
    \hspace*{0mm}
    	\begin{minipage}[t]{\imwidth\linewidth}
    		\centering
			\begin{tikzpicture}[spy using outlines={green,magnification=\ssmag,size=\ssizz},inner sep=0]
				\node {\includegraphics[width=\linewidth,trim={0 0 0 0},clip,cframe=black .005mm]{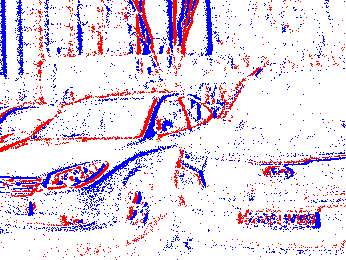}};
				\spy on \sxxxxl in node [left] at \ssyysr;
				\spy[red] on \sxxxxrl in node [left,red] at \ssyys;
				\end{tikzpicture}
			(b) Events \\
            (PSNR, SSIM)
            \vspace{0.5em}
                
			\begin{tikzpicture}[spy using outlines={green,magnification=\ssmag,size=\ssizz},inner sep=0]
				\node {\includegraphics[width=\linewidth]{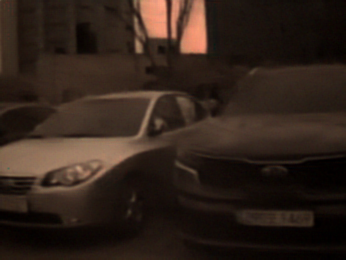}};
				\spy on \sxxxl in node [left] at \ssyysr;
				\spy[red] on \sxxxrl in node [left,red] at \ssyys;
				\end{tikzpicture}
			(g) w/o $\mathcal{L}_{B\text{-}E}$ \\
            (31.29, 0.9235) \vspace{0.3em}
    	\end{minipage}%
    \hspace*{0mm}
    	\begin{minipage}[t]{\imwidth\linewidth}
    		\centering
			\begin{tikzpicture}[spy using outlines={green,magnification=\ssmag,size=\ssizz},inner sep=0]
				\node {\includegraphics[width=\linewidth]{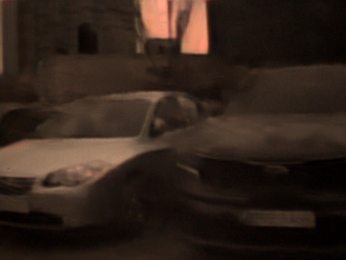}};
				\spy on \sxxxxl in node [left] at \ssyysr;
				\spy[red] on \sxxxxrl in node [left,red] at \ssyys;
				\end{tikzpicture}
			(c) w/ $\mathcal{L}_{S\text{-}B}$ \\
            (27.34, 0.8575)
            \vspace{0.5em}
                
			\begin{tikzpicture}[spy using outlines={green,magnification=\ssmag,size=\ssizz},inner sep=0]
				\node {\includegraphics[width=\linewidth]{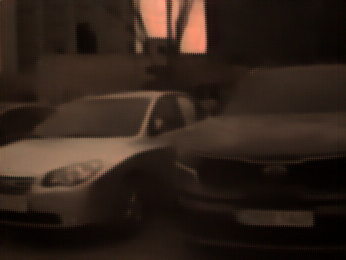}};
				\spy on \sxxxl in node [left] at \ssyysr;
				\spy[red] on \sxxxrl in node [left,red] at \ssyys;
				\end{tikzpicture}
			(h) w/o $\mathcal{L}_{B\text{-}B}$ \\
            (29.57, 0.8786) \vspace{0.3em}
    	\end{minipage}%
    \hspace*{0mm}
		\begin{minipage}[t]{\imwidth\linewidth}
    		\centering
			\begin{tikzpicture}[spy using outlines={green,magnification=\ssmag,size=\ssizz},inner sep=0]
				\node {\includegraphics[width=\linewidth]{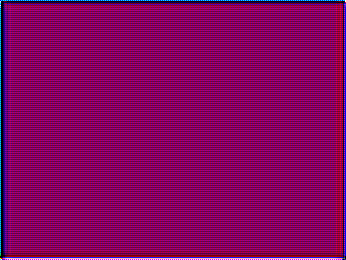}};
				\spy on \sxxxxl in node [left] at \ssyysr;
				\spy[red] on \sxxxxrl in node [left,red] at \ssyys;
				\end{tikzpicture}
			(d) w/ $\mathcal{L}_{B\text{-}E}$ \\
            (9.71, 0.0280)
            \vspace{0.5em}
                
			\begin{tikzpicture}[spy using outlines={green,magnification=\ssmag,size=\ssizz},inner sep=0]
				\node {\includegraphics[width=\linewidth]{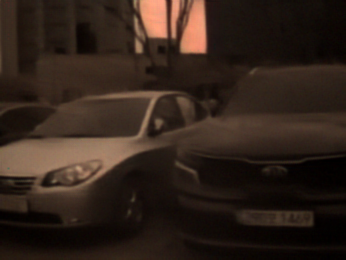}};
				\spy on \sxxxl in node [left] at \ssyysr;
				\spy[red] on \sxxxrl in node [left,red] at \ssyys;
				\end{tikzpicture}
			(i) w/ all \\
            (\textbf{32.79}, \textbf{0.9317}) \vspace{0.3em}
    	\end{minipage}%
    \hspace*{0mm}
    	\begin{minipage}[t]{\imwidth\linewidth}
    		\centering
			\begin{tikzpicture}[spy using outlines={green,magnification=\ssmag,size=\ssizz},inner sep=0]
				\node {\includegraphics[width=\linewidth]{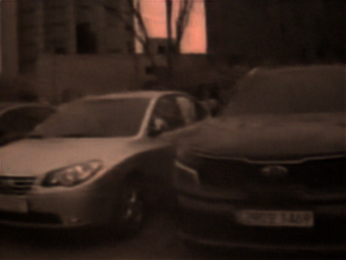}};
				\spy on \sxxxxl in node [left] at \ssyysr;
				\spy[red] on \sxxxxrl in node [left,red] at \ssyys;
				\end{tikzpicture}
			(e) w/ $\mathcal{L}_{B\text{-}B}$ \\
            (30.43, 0.9164)
            \vspace{0.5em}
                
			\begin{tikzpicture}[spy using outlines={green,magnification=\ssmag,size=\ssizz},inner sep=0]
				\node {\includegraphics[width=\linewidth]{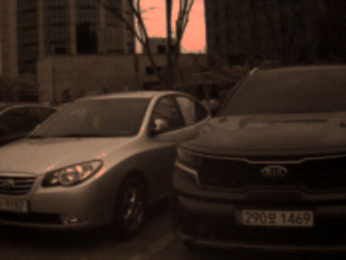}};
				\spy on \sxxxl in node [left] at \ssyysr;
				\spy[red] on \sxxxrl in node [left,red] at \ssyys;
				\end{tikzpicture}
			(j) GT \\
            (Inf., 1.0000) \vspace{0.3em}
    	\end{minipage}%
    \centering
	\caption{Qualitative ablations of each self-supervised loss on the MD task. Metric performance (\ie, PSNR$\uparrow$/SSIM$\uparrow$) on estimating the latent sharp image over the ColorDVS dataset.}
	\label{fig_ablation_loss}
\end{figure*}

\subsubsection{Performance of Blurry Video Enhancement}
For the task of Blurry Video Enhancement (BVE), we use 13 consecutive frames as input for the REDS dataset and 25 consecutive frames for the ColorDVS dataset. In the REDS dataset, we synthesize two blurry images using the first and last 5 frames, while in the ColorDVS dataset, we use the initial and final 9 frames. Performance assessment involves restoring 3 sharp frames between the two blurry images for both datasets. To ensure consistency with the 3-frame restoration of the REDS dataset, one frame is skipped in the ColorDVS dataset. We compare our EVDI and EVDI++ to state-of-the-art frame-based methods, including Jin~\cite{jin2018learning}, Bin~\cite{shen2020blurry}, and DeMFI~\cite{oh2022demfi}, and event-based approaches, including EDI~\cite{edi_pan2019bringing} and REFID~\cite{sun2023event}. The quantitative results are presented in \cref{tab:blur_interp}, where our proposed EVDI and \myname\ outperform both the frame- and event-based BVE approaches. According to the qualitative results in \cref{fig:exp_blurry_interp}, we can observe that both EVDI and \myname\ exhibit clearer details of the hair in the top two rows and the sharper edge of the window in the bottom two rows compared to other frame- and event-based methods. Meanwhile, EVDI++ achieves superior quantitative and qualitative performance compared to EVDI, demonstrating the effectiveness of the proposed key modules, including LDR and APF. Moreover, in contrast to existing approaches that are limited to recovering only a single or a restricted number of latent sharp frames, our proposed EVDI++ demonstrates the capability to restore latent sharp frames at arbitrary timestamps within $\mathcal{T}_{i+1}^{i}$. The video reconstruction demonstration is available on our project page: \url{https://bestrivenzc.github.io/EVDI-plus-plus/}.

\subsection{Ablation Study} \label{exp_ablation}
In this subsection, we perform ablation studies on the self-supervised losses presented in \cref{ablation_loss} and the network architecture, including Exposure-Transferred Reconstruction (LDR) and Adaptive Parameter-free Fusion (APF), introduced in \cref{ablation_network}. Additionally, we conducted comprehensive ablation experiments for the three self-supervised loss functions and the network architecture using the ColorDVS dataset. This involved retraining the network without the corresponding ablation items on the ColorDVS training set and evaluating the performance on the testing set.


\def\imwidth{0.33}
\def\cimwid{0.14}

\def\zuoxia{(-0.3,-0.9)}
\def\youshang{(0.2,0.45)}

\def\ssyy{(-0.8,-0.85)}
\def\ssizz{0.5cm}
\def\sswidth{0.245\textwidth}
\def\ssmag{3}
\def\scc{(2.12,1.4)}

\begin{figure}[!htb]
\footnotesize
	\centering
\begin{minipage}[t]{\imwidth\linewidth}
    		\centering
			\begin{tikzpicture}[spy using outlines={rectangle,green,magnification=\ssmag,size=\ssizz},inner sep=0]
				\node {\includegraphics[width=\linewidth]{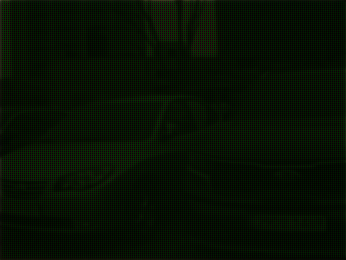}};
				\end{tikzpicture}
            $\bar{L}_{i}(m)$ 
            \\
                        \vspace{.2mm}
            \begin{tikzpicture}[spy using outlines={green,magnification=\ssmag,size=\ssizz},inner sep=0]
				\node {\includegraphics[width=\linewidth]{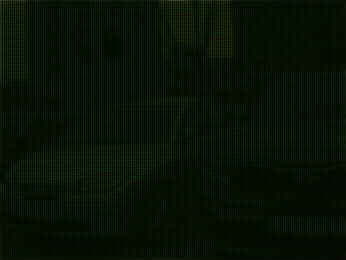}};
				\end{tikzpicture}
			$\bar{L}_{i+1}(m)$
\end{minipage}%
    \hfill
\begin{minipage}[t]{\imwidth\linewidth}
    		\centering
    	    \begin{tikzpicture}[spy using outlines={green,magnification=\ssmag,size=\ssizz},inner sep=0]
				\node {\includegraphics[width=\linewidth]{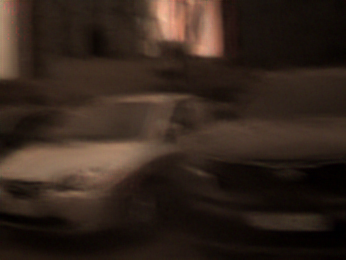}};
				\end{tikzpicture}
            $\bar{B}_{i}$
   \\
      \vspace{.5mm}
   \begin{tikzpicture}[spy using outlines={green,magnification=\ssmag,size=\ssizz},inner sep=0]
				\node {\includegraphics[width=\linewidth]{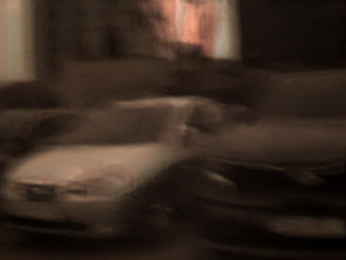}};
				\end{tikzpicture}
			$\bar{B}_{i+1}$
\end{minipage}%
    \hfill
\begin{minipage}[t]{\imwidth\linewidth}
    		\centering
    	    \begin{tikzpicture}[spy using outlines={green,magnification=\ssmag,size=\ssizz},inner sep=0]
				\node {\includegraphics[width=\linewidth]{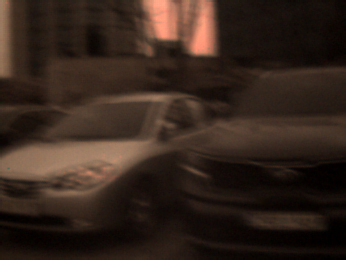}};
				\end{tikzpicture}
                $B_{i}$
   \\
      \vspace{.5mm}
            \begin{tikzpicture}[spy using outlines={green,magnification=\ssmag,size=\ssizz},inner sep=0]
				\node {\includegraphics[width=\linewidth]{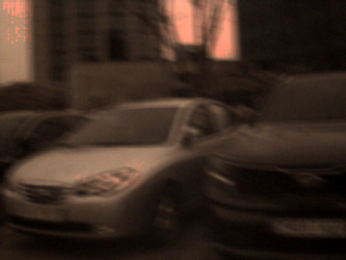}};
				\end{tikzpicture}
			         $B_{i+1}$ 
    	\end{minipage}%

\caption{Qualitative results for the MD task in Case (5) of \cref{tab:ablation_loss}, comprising latent sharp images $\bar{L}_{i}(m)$ and $\bar{L}_{i+1}(m)$, re-blurred images $\bar{B}_i$ and $\bar{B}_{i+1}$, and the corresponding input blurry images $B_i$ and $B_{i+1}$.}
  \vspace{-1em}
    \label{Ablation_wo_bs}
  \end{figure}

\begin{table}[hbpt]
    \centering
    \caption{Ablation study of supervisions on the ColorDVS dataset with real-world events. \textbf{Bold} and \underline{underlined} numbers represent the best and the second-best performance.}
    \resizebox{\linewidth}{!}{
    \begin{tabular}{c|cccccccccc}
\hline 
        \multirow{2}{*}{Models} & \multirow{2}{*}{$\mathcal{L}_{S-B}$} & \multirow{2}{*}{$\mathcal{L}_{B-E}$} & \multirow{2}{*}{$\mathcal{L}_{B-B}$} & \multicolumn{3}{c}{PSNR$\uparrow$/SSIM$\uparrow$} \\ \cline{5-7}
                &  &  &  & MD & FI & BVE \\ \hline
            \#0    &  \checkmark      &                   &                   &   28.84/0.8007    &   31.78/0.8728    &   29.98/0.8126      \\ 
             \#1    &  &           \checkmark        &                   &       10.47/0.0193 &  13.29/0.1846     & 7.89/0.0367    \\
             \#2    &  &                   &        \checkmark           &      31.39/0.8762 &    32.13/0.8772   &  30.50/0.8565  \\
          \#3 &\checkmark       &      \checkmark             &                   & 30.04/0.8354      &  32.15/0.8766     & 30.16/0.8278      \\
           \#4   &\checkmark    &                   &     \checkmark              &  \underline{31.57}/\underline{0.8774}     &  \underline{32.24}/\underline{0.8775}     &  \underline{31.17}/\underline{0.8576} \\
           \#5   &    &          \checkmark         &         \checkmark          &    12.72/0.1203   &  12.79/0.1527     &  12.73/0.1471   \\
           \#6 &   \checkmark   &         \checkmark          &       \checkmark            &    \textbf{32.05}/\textbf{0.8853}    &   \textbf{32.83}/\textbf{0.8883}     &   \textbf{31.64}/\textbf{0.8678}  \\ \hline
        
    \end{tabular}
    }
    \label{tab:ablation_loss}
\end{table}

\subsubsection{Necessity of Loss Combination}\label{ablation_loss}

The self-supervised losses proposed in our EVDI++ consist of Sharp-Blurry $\mathcal{L}_{S\text{-}B}$, Blurry-Event $\mathcal{L}_{B\text{-}E}$, and Blurry-Blurry $\mathcal{L}_{B\text{-}B}$ losses. The qualitative and quantitative results on the ColorDVS dataset are presented in \cref{fig_ablation_loss} and \cref{tab:ablation_loss}, respectively.

\noindent \textbf{Sharp-Blurry Loss.} 
The sharp-blurry loss $\mathcal{L}_{S\text{-}B}$ is devised to ensure brightness consistency between the predicted latent sharp image with the input blurry image. In \cref{fig_ablation_loss} (c), using only $\mathcal{L}_{S\text{-}B}$ maintains brightness consistency with the input blurry image (a), showcasing its effectiveness. Qualitative results in \cref{fig_ablation_loss} comparing case (d) with (h), and (m) with (i) further affirm that $\mathcal{L}_{S\text{-}B}$ contributes to maintaining brightness consistency. Moreover, Cases (1) and (3), (2) and (4), (5) and (6) in \cref{tab:ablation_loss} emphasize average quantitative improvements PSNR (13.02 dB) and SSIM (0.5271) across the three tasks attributable to $\mathcal{L}_{S\text{-}B}$.     

\def\imwidth{0.33}
\def\cimwid{0.14}

\def\zuoxia{(-0.3,-0.9)}
\def\youshang{(0.2,0.45)}

\def\ssyy{(-0.8,-0.85)}
\def\ssizz{0.5cm}
\def\sswidth{0.245\textwidth}
\def\ssmag{3}
\def\scc{(2.12,1.4)}

\begin{figure}[!htb]
\footnotesize
	\centering
\begin{minipage}[t]{\imwidth\linewidth}
    		\centering
			\begin{tikzpicture}[spy using outlines={rectangle,green,magnification=\ssmag,size=\ssizz},inner sep=0]
				\node {\includegraphics[width=\linewidth]{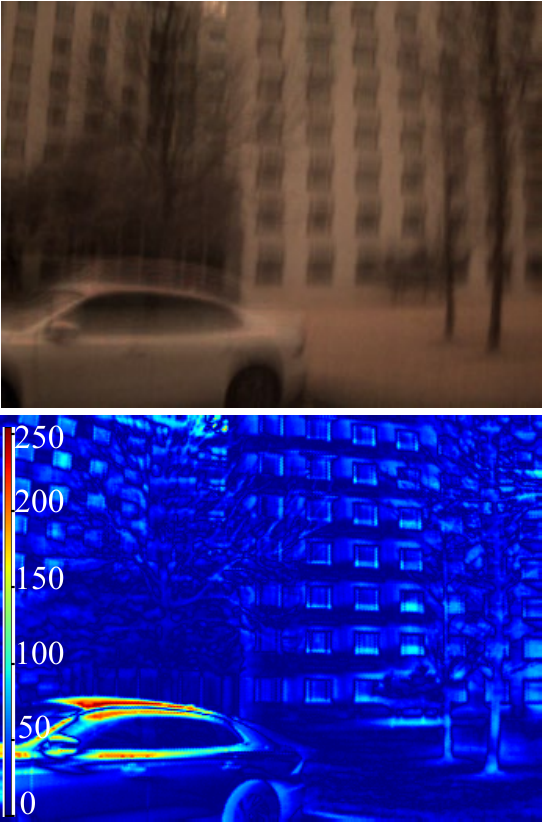}};
				\end{tikzpicture}

			(a) Blurry image\\
			(PSNR, SSIM) \vspace{0.5em}
\end{minipage}%
    \hfill
\begin{minipage}[t]{\imwidth\linewidth}
    		\centering
    	    \begin{tikzpicture}[spy using outlines={green,magnification=\ssmag,size=\ssizz},inner sep=0]
				\node {\includegraphics[width=\linewidth]{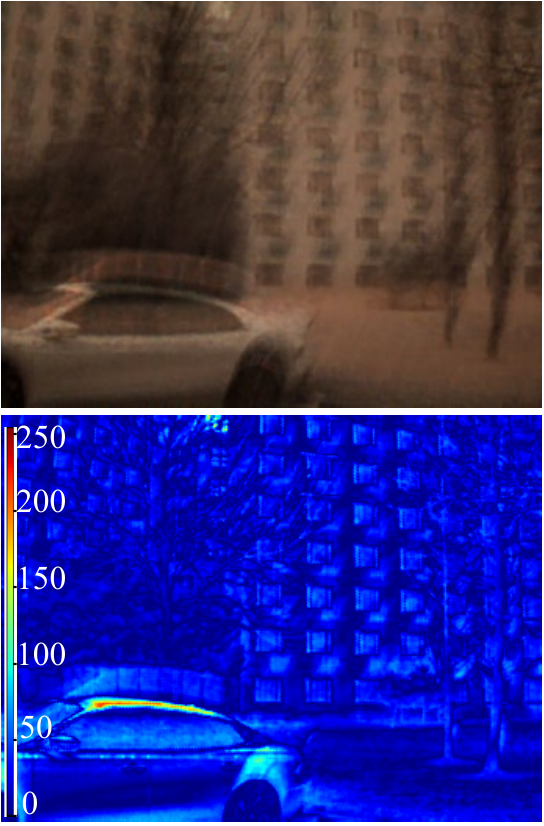}};
				\end{tikzpicture}

			(b) w/o ETR+APF 
   \\
			(22.44, 0.6736)\vspace{0.5em}
\end{minipage}%
    \hfill
\begin{minipage}[t]{\imwidth\linewidth}
    		\centering
    	    \begin{tikzpicture}[spy using outlines={green,magnification=\ssmag,size=\ssizz},inner sep=0]
				\node {\includegraphics[width=\linewidth]{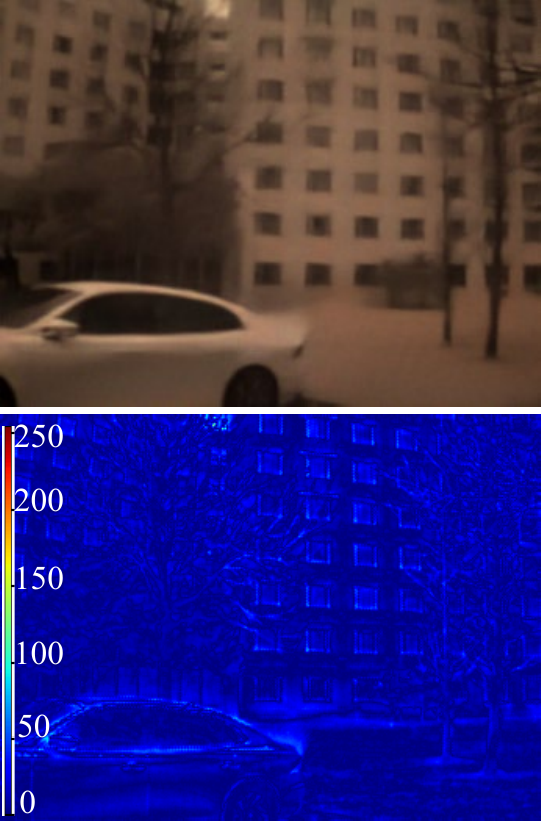}};
				\end{tikzpicture}

			(c) w/o APF 
   \\
	(28.77, 0.8121)\vspace{0.5em}
    	\end{minipage}%
    \hfill
\begin{minipage}[t]{\imwidth\linewidth}
			\centering
			\begin{tikzpicture}[spy using outlines={green,magnification=\ssmag,size=\ssizz},inner sep=0]
				\node {\includegraphics[width=\linewidth]{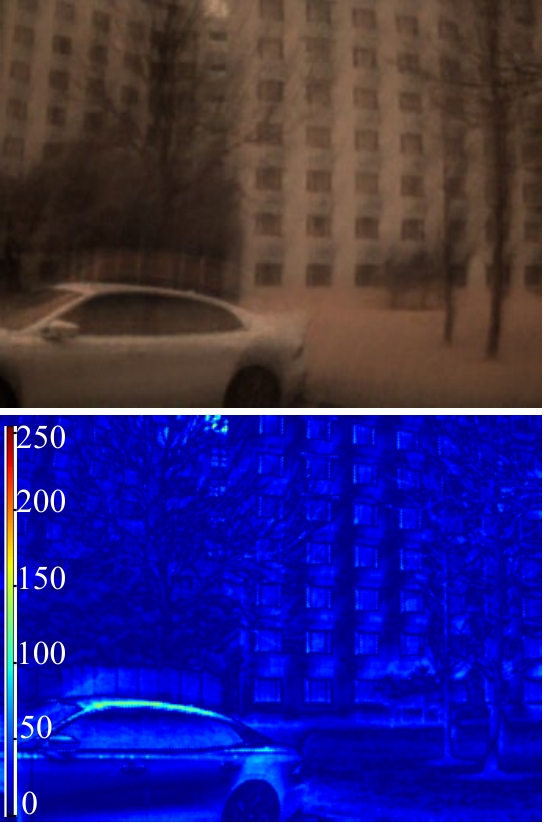}};
				\end{tikzpicture}

			(d) w/o ETR 
   \\
	(24.98, 0.7685)\vspace{0.5em}
\end{minipage}%
   \hfill
\begin{minipage}[t]{\imwidth\linewidth}
    		\centering
    	    \begin{tikzpicture}[spy using outlines={green,magnification=\ssmag,size=\ssizz},inner sep=0]
				\node {\includegraphics[width=\linewidth]{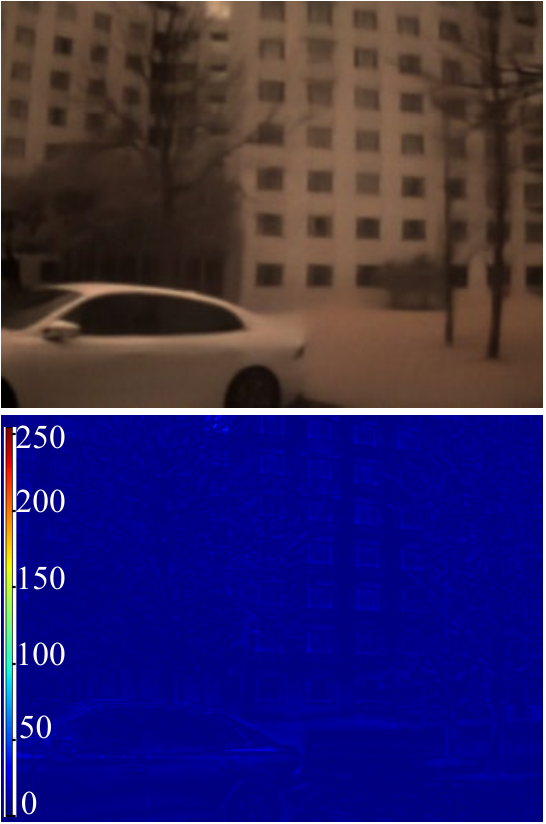}};
				\end{tikzpicture}

			(e) w/ all
   \\
			(\textbf{29.89}, \textbf{0.8362)}\vspace{0.5em}
    	\end{minipage}%
     \hfill
\begin{minipage}[t]{\imwidth\linewidth}
    		\centering
    	    \begin{tikzpicture}[spy using outlines={green,magnification=\ssmag,size=\ssizz},inner sep=0]
				\node {\includegraphics[width=\linewidth]{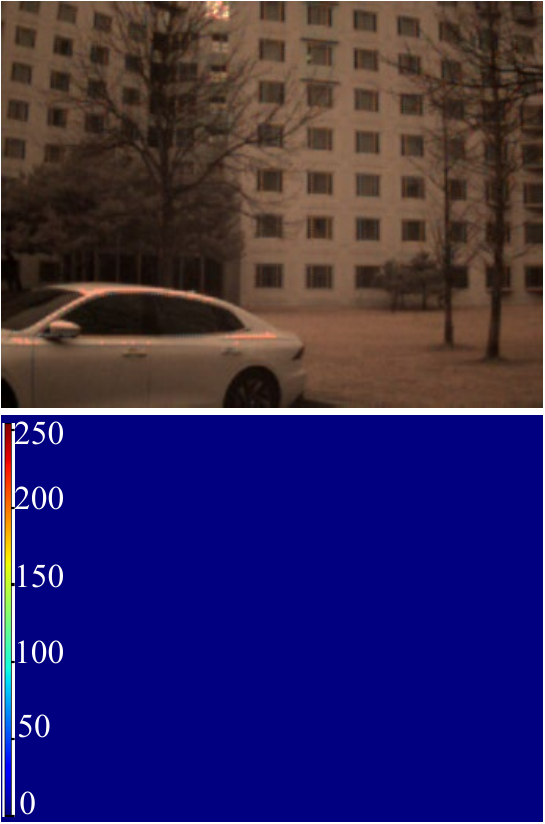}};
				\end{tikzpicture}
			(f) GT
   \\
			(Inf., 1.000)\vspace{0.5em}
    	\end{minipage}%

\caption{Ablations of network modules of EVDI++ for the BVE task. Qualitative (recovered latent sharp images and corresponding absolute errors) and quantitative results are given over the ColorDVS dataset.}
\vspace{-3mm}
\label{fig:ablation_moudle}
\end{figure}

\noindent \textbf{Blurry-Event Loss.} Lacking the constraint of brightness consistency, the model fails to recover latent results with appropriate brightness when relying solely on the blurry-event loss $\mathcal{L}_{B\text{-}E}$, as shown in \cref{fig_ablation_loss} (d) and case (1) of \cref{tab:ablation_loss}. Comparisons between cases (g) and (i), (c) and (h) in \cref{fig_ablation_loss}, cases (0) and (3), and (4) and (6) in \cref{tab:ablation_loss} reveal that $\mathcal{L}_{B\text{-}E}$ plays a crucial role in enhancing image sharpness since $\mathcal{L}_{B\text{-}E}$ can deal with motion ambiguity by gaining supervision from reference frames and events. Nevertheless, an interesting phenomenon emerges from the comparison between cases (e) and (m) in \cref{fig_ablation_loss}, and cases (2) and (5) in \cref{tab:ablation_loss}, indicating that the inclusion of $\mathcal{L}_{B\text{-}E}$ results in a deterioration of performance. This phenomenon arises from the distinction between $\mathcal{L}_{S\text{-}B}$ and $\mathcal{L}_{B\text{-}B}$. Both losses enforce brightness consistency, but $\mathcal{L}_{S\text{-}B}$ achieves this directly for the sharp latent image $\bar{L}(m)$. On the other hand, $\mathcal{L}_{B\text{-}B}$ employs cross-supervision to simultaneously maintain brightness consistency for the reconstructed blurry image $\bar{B}$ and exploit the complementary sharpness information embedded in the adjacent real blurry frames $B_i$ and $B_{i+1}$, as detailed in \cref{Blur_Blur_evdi++}. Consequently, combining $\mathcal{L}_{B\text{-}B}$ and $\mathcal{L}_{B\text{-}E}$ result in both the reconstructed latent sharp images $\bar{L}_{i}(m)$ and $\bar{L}_{i+1}(m)$ having values close to 0 to satisfy \cref{be_LDR_L}, while the reconstructed blurry images $\bar{B}_{i}$ and $\bar{B}_{i+1}$ maintain brightness consistency to fulfill \cref{Blur_Blur_evdi++}, as illustrated in \cref{Ablation_wo_bs}.

\noindent \textbf{Blurry-Blurry Loss.} Solely employing the blurry-blurry loss $\mathcal{L}_{B\text{-}B}$ show favorable quantitative (Case (2) of \cref{tab:ablation_loss}) and qualitative (\cref{fig_ablation_loss} (e)) results, preserving not only brightness consistency in the reconstructed latent image but also restoring sharper texture details. The deletion of the $\mathcal{L}_{B\text{-}B}$ (comparing case (c) with (g), and (h) with (i)) brings performance degeneration, \ie, 2.37 dB and 0.0633 in terms of PSNR and SSIM, and corresponding visualization results shown in \cref{fig_ablation_loss} (c) and (h) suffer severe artifacts, which validates the effectiveness of the $\mathcal{L}_{B\text{-}B}$. 

\subsubsection{Network Architecture}\label{ablation_network}
The proposed network architecture in our EVDI++ is composed of a Learning-based Division Reconstruction module (LDR) and an Adaptive Parameter-free Fusion (APF) module. Ablation studies are conducted on the ColorDVS dataset with real-world events, where four different experiments are implemented to analyze the effectiveness of each module. Quantitative and qualitative results are shown in \cref{ablation_module} and \cref{fig:ablation_moudle}.

\begin{table}[htb]
\caption{Ablation study of the LDR and APF modules on the ColorDVS dataset. \textbf{Bold} numbers represent the best performance.}
\centering
\begin{tabular}{c|cccccc}

\hline
\multirow{2}{*}{Case} & \multirow{2}{*}{LDR} & \multirow{2}{*}{APF} & \multicolumn{3}{c}{PSNR$\uparrow$/SSIM$\uparrow$} \\ \cline{4-6}
              &  &  &  MD & FI & BVE \\ \hline
      \#0 &      &     & 29.53/0.8667 & 23.47/0.6990
      & 25.28/0.7792
      \\
\#1   & \checkmark      &           & \textbf{32.05}/\textbf{0.8853} & 30.59/0.8663
& 30.84/0.8564 
\\
 \#2   &    &\checkmark           & 29.53/0.8667 & 25.61/0.7659 & 27.24/0.8163 \\
\#3   & \checkmark     & \checkmark          & \textbf{32.05}/\textbf{0.8853}  &  \textbf{32.83}/\textbf{0.8883} & \textbf{31.64}/\textbf{0.8676} \\ \hline
\end{tabular}
\label{ablation_module}
\end{table}

\noindent \textbf{Learning-based Division Reconstruction.} First, we eliminate the LDR module and directly estimate the latent sharp images, $\bar{L}(m)$, using \cref{blurry_latent_event}. This results in a significant decline in both quantitative and qualitative performance. Specifically, the removal of the LDR module, as shown in cases (0) and (2) in \cref{ablation_module}, results in a degradation of 6.36 dB in PSNR and 0.0589 in SSIM. Qualitatively, the visual results in \cref{fig:ablation_moudle} (b) and (d) without the LDR module exhibit pronounced distortions, while cases (c) and (e), incorporating the LDR module, produce reconstructions with sharper details, confirming its effectiveness.       

\noindent \textbf{Adaptive Parameter-free Fusion.} Moreover, the APF module is designed to fuse the reconstructed latent images $\bar{L}_{i}(m)$ and $\bar{L}_{i+1}(m)$, generating refined final results $\tilde{L}(m)$ when $f\in \mathcal{T}_{i\rightarrow{i+1}}$. It directly outputs $\bar{L}_{i}(m)$ and $\bar{L}_{i+1}(m)$ when $f\in \mathcal{T}_{i}$ or $f\in \mathcal{T}_{i+1}$ to alleviate cumulative errors. Hence, the quantitative results for the MD task are identical between cases (0) and (2), as well as between cases (1) and (3), as shown in \cref{ablation_module}. Qualitative results in \cref{fig:ablation_moudle} comparing (b) with (d), and (c) with (e) demonstrate that EVDI++ with the APF module gives sharper results than the framework without it, validating the effectiveness of the APF module. Similarly, a consistent conclusion can be derived from the quantitative comparison experiments, \ie, (0) \vs ~(2), and (1) \vs ~(3), presented in \cref{ablation_module}. \re{Note that for motion deblurring tasks where the target timestamp lies within an original exposure interval, the APF module degenerates to a simple identity mapping and thus does not alter the reconstruction result.}

\def\ssxxsone{(-0.9, -0.28)} 
\def\ssevdisone{(-0.95, -0.15)}
\def\ssyysone{(2.08, 0.79)} 

\def\ssxxstwo{(-0.55,-0.90)} 
\def\ssyystwo{(2.08,-0.79)} 

\def\ssizz{1cm} 
\def\sswidth{0.235\textwidth} 
\def\scc{(1.70,-1.15)}
\def\sccone{(-2.12,-1.45)}

\def\ssizzone{1.5cm} 
\def\sswidthone{0.235\textwidth} 
\def\ssmagone{3}

\begin{figure*}[!thb]
	\centering

    \begin{tikzpicture}[spy using outlines={green,magnification=\ssmagone,size=\ssizzone},inner sep=0]
		\node {\includegraphics[width=\sswidth]{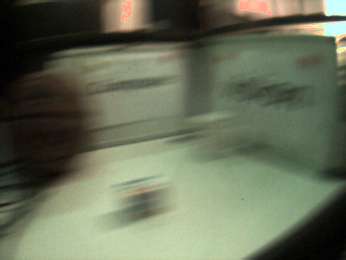}};
		\spy on \ssxxstwo in node [left] at \ssyystwo;
		\node [anchor=west, font=\newimgfont] at \sccone {\textcolor{yellow}{\bf Start Frame}};
	\end{tikzpicture} 
    \begin{tikzpicture}[spy using outlines={green,magnification=\ssmagone,size=\ssizzone},inner sep=0]
		\node {\includegraphics[width=\sswidth]{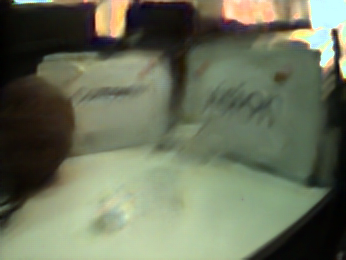}};
		\spy on \ssxxstwo in node [left] at \ssyystwo;
		\node [anchor=west, font=\newimgfont] at \sccone {\textcolor{yellow}{\bf Jin}};
	\end{tikzpicture} 
	\begin{tikzpicture}[spy using outlines={green,magnification=\ssmagone,size=\ssizzone},inner sep=0]
		\node {\includegraphics[width=\sswidth]{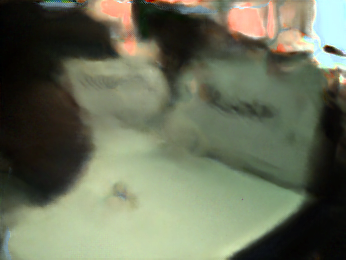}};
		\spy on \ssxxstwo in node [left] at \ssyystwo;
		\node [anchor=west, font=\newimgfont] at \sccone {\textcolor{yellow}{\bf DeMFI}};
	\end{tikzpicture} 
	\begin{tikzpicture}[spy using outlines={green,magnification=\ssmagone,size=\ssizzone},inner sep=0]
		\node {\includegraphics[width=\sswidth]{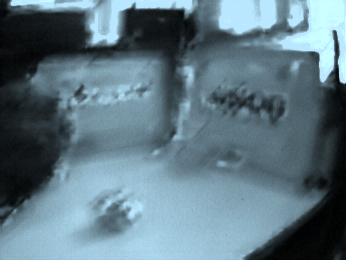}};
		\spy on \ssxxstwo in node [left] at \ssyystwo;
		\node [anchor=west, font=\newimgfont] at \sccone {\textcolor{yellow}{\bf LEDVDI}};
	\end{tikzpicture}

 \vspace{.5mm}

    \begin{tikzpicture}[spy using outlines={green,magnification=\ssmagone,size=\ssizzone},inner sep=0]
		\node {\includegraphics[width=\sswidth]{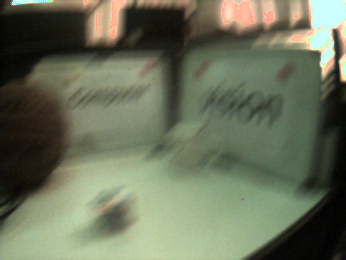}};
		\spy on \ssxxstwo in node [left] at \ssyystwo;
		\node [anchor=west, font=\newimgfont] at \sccone {\textcolor{yellow}{\bf End Frame}};
	\end{tikzpicture} 
    \begin{tikzpicture}[spy using outlines={green,magnification=\ssmagone,size=\ssizzone},inner sep=0]
		\node {\includegraphics[width=\sswidth]{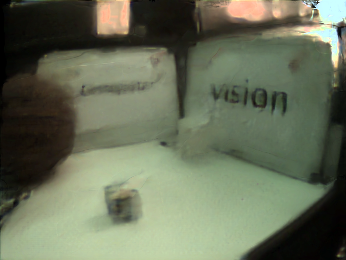}};
		\spy on \ssxxstwo in node [left] at \ssyystwo;
		\node [anchor=west, font=\newimgfont] at \sccone {\textcolor{yellow}{\bf TimeLens}};
	\end{tikzpicture} 
	\begin{tikzpicture}[spy using outlines={green,magnification=\ssmagone,size=\ssizzone},inner sep=0]
		\node {\includegraphics[width=\sswidth]{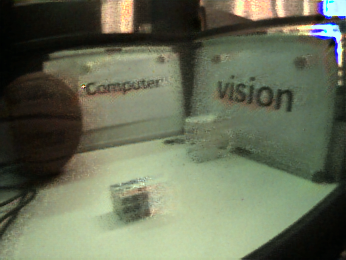}};
		\spy on (-0.50,-0.88) in node [left] at \ssyystwo;
		\node [anchor=west, font=\newimgfont] at \sccone {\textcolor{yellow}{\bf EVDI}};
	\end{tikzpicture} 
	\begin{tikzpicture}[spy using outlines={green,magnification=\ssmagone,size=\ssizzone},inner sep=0]
		\node {\includegraphics[width=\sswidth]{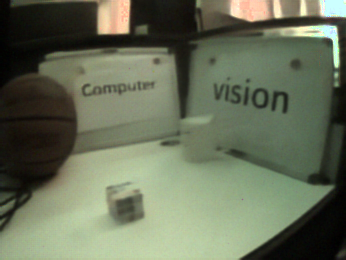}};
		\spy on \ssxxstwo in node [left] at \ssyystwo;
		\node [anchor=west, font=\newimgfont] at \sccone {\textcolor{yellow}{\bf EVDI++}};
	\end{tikzpicture}
	\caption{Qualitative results on the ColorRBE datasets with real-world blurry images and events}
    \vspace{-1em}
	\label{fig:exp_hrbe}
\end{figure*}

\subsection{Evaluation on the ColorRBE dataset}\label{exp_hrbe}
To demonstrate adaptivity to the real-world scenario of our proposed EVDI++, we conduct comparison experiments on the ColorRBE dataset with real-world blurry frames and corresponding events. Our method is compared against state-of-the-art frame-based approaches, including Jin~\cite{jin2018learning} and DeMFI~\cite{oh2022demfi}, and event-based methods, including LEDVDI~\cite{ledvdi_lin2020learning} and TimeLens~\cite{tulyakov2021time}. The qualitative results are shown in \cref{fig:exp_hrbe}. The comparison methods take the consecutive real blurry frames as input and reconstruct the latent sharp image at the middle timestamps of the entire exposure interval. Jin's work employs a cascaded scheme for deblurring and interpolation, which tends to propagate the deblurring error to the interpolation stage. The performance of TimeLens is highly dependent on the quality of reference frames, and DeMFI and LEDVDI often face performance drops when inferring on real-world datasets owing to data inconsistency. Our proposed EVDI++ tackles the inconsistency problem by learning the target scenarios with the self-supervised framework, thus producing better results. Furthermore, EVDI++ demonstrates superior generalization performance compared to its previous version EVDI, validating the effectiveness of the LDR, APF, and refined self-supervised losses. The video reconstruction demonstration on the ColorRBE dataset is also available on our project page: \url{https://bestrivenzc.github.io/EVDI-plus-plus/}.

\begin{table}[!t]
\centering
\scriptsize
\caption{Runtime comparison (in milliseconds) between the proposed EVDI++ and other representative algorithms across the tasks of MD, FI, and BVE by using 1 NVIDIA TITAN RTX 3090 GPU.}
\vspace{-3mm}
\begin{tabular}{lccccccccccc}
\hline
\multirow{2}{*}{Methods} & \multirow{2}{*}{Task} & \multirow{2}{*}{Events} & \multicolumn{3}{c}{Resolution}
\\  \cline{4-6} & & & \multicolumn{1}{c}{346$\times$260} & \multicolumn{1}{c}{640$\times$480} & \multicolumn{1}{c}{1280$\times$720} \\
\hline
LEVS  & MD & \XSolidBrush & 153.0 & 202.8 & 576.2
\\
MPRNet  & MD & \XSolidBrush & 127.6 & 425.6 & 1242.3
\\
RED-Net  & MD & \Checkmark & 42.9 & 101.3 & 329.2
\\
EF-Net  & MD & \Checkmark & 36.1 & 112.3 & 314.8
\\
RIFE  & FI & \XSolidBrush & 15.8 & 16.6 & 20.4
\\
EMA  & FI & \XSolidBrush & 296.7 & 728.5 & 2659.6
\\
UnSuperS  & FI & \XSolidBrush & 53.2 & 123.5 & 371.3
\\
Timelens  & FI & \Checkmark & 57.8 & 104.4 & 205.2
\\
REFID  & FI & \Checkmark & 71.3 & 258.6 & 649.0
\\
Jin  & BVE & \XSolidBrush & 1273.2 & 5382.0 & 17327.8 
\\
Bin  & BVE & \XSolidBrush & 272.5 & 547.3 & 1587.5
\\
DeMFI  & BVE & \XSolidBrush & 167.2 & 526.3 & 1497.4
\\
EVDI  & MD,BVE & \Checkmark & 18.1 & 56.8 & 141.1
\\
EVDI++  & MD,FI,BVE & \Checkmark & 27.1 & 46.5 & 89.5
\\
\hline
\end{tabular}
\label{tab:runtime}
\vspace{-5mm}
\end{table}

\subsection{Complexity to Performance Analysis} \label{sec_complexity}
We further evaluate the relationship between the complexity and performance of our EVDI++ and state-of-the-art motion deblurring, frame interpolation, and blurry video enhancement methods on the synthetic REDS dataset, as shown in \cref{Overview} (b). In all three tasks, our EVDI++ demonstrates substantial advantages in terms of both effectiveness and model complexity. Furthermore, EVDI++ outperforms its previous version EVDI in both performance and model complexity across the MD and BVE tasks, which shows the effectiveness of the exposure-transferred reconstruction, adaptive parameter-free fusion modules, and the improved self-supervised losses, \ie, $\mathcal{L}_{S\text{-}B}$, $\mathcal{L}_{B\text{-}E}$, and $\mathcal{L}_{B\text{-}B}$, presented in EVDI++. Furthermore, we provide detailed comparative statistics on the inference times of different algorithms across various resolutions, as presented in \cref{tab:runtime}. Notably, our proposed EVDI++ exhibits significant advantages in inference time across all resolutions, except when compared to the real-time frame interpolation method RIFE~\cite{huang2022real}.

\begin{figure}[!t]
    \centering
    \includegraphics[width=0.99\linewidth]{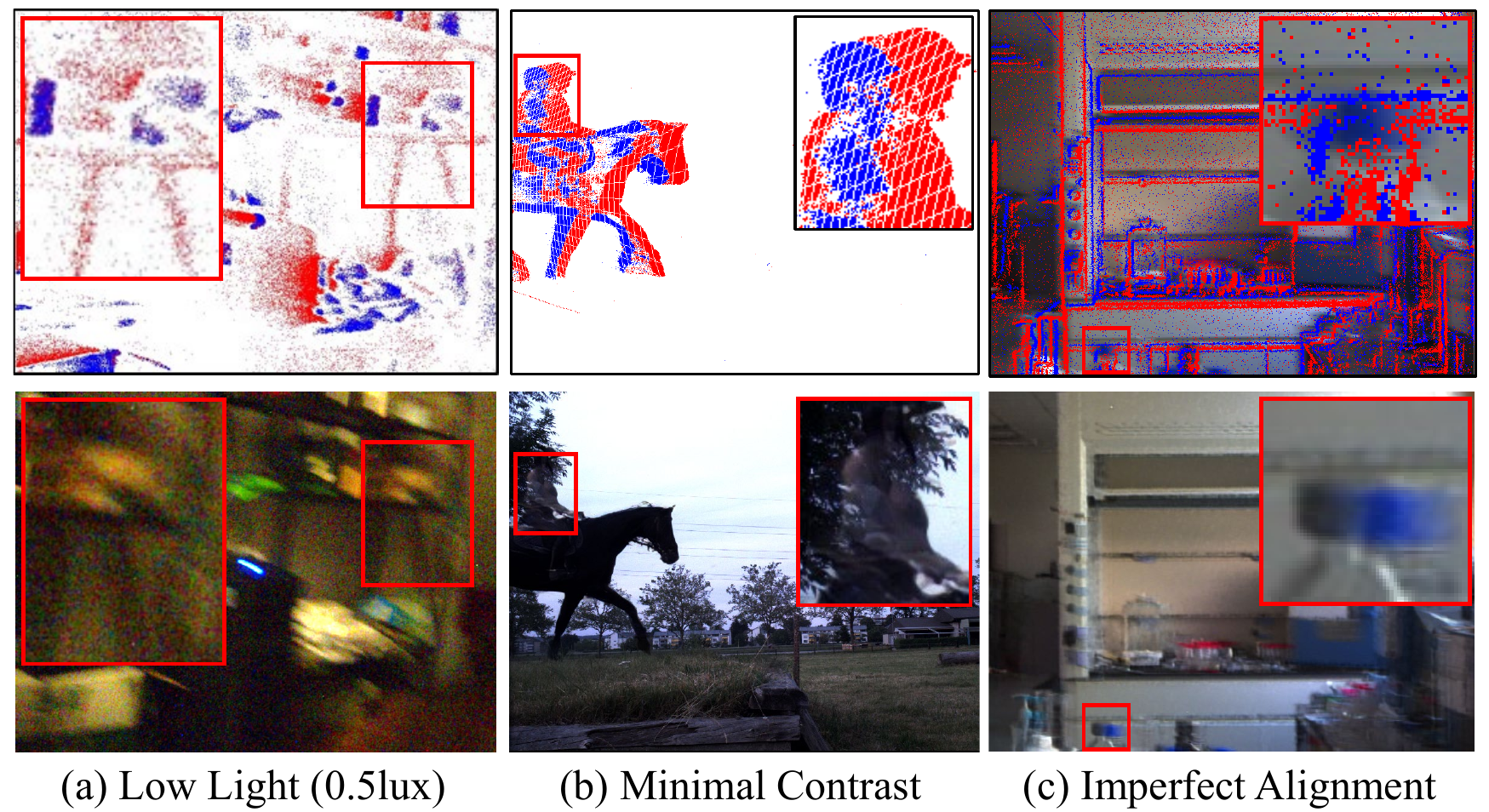}
    \caption{Examples of failure cases.}
    \label{limitation}
\end{figure}

\subsection{Limitations and Future Works} \label{Limitations}
As shown in \cref{limitation}, we identify failure cases to highlight the limitations of EVDI++. The performance critically depends on event stream fidelity, as accurate estimation of the double integrals $E(f,\mathcal{T}_i)$ and $E(f,\mathcal{T}_{i+1})$ requires reliable event data. In challenging scenarios such as low-light conditions and minimal foreground-background contrast, degraded event generation leads to excessive noise and sparse events, significantly compromising reconstruction quality (\cref{limitation} (a) and (b)). Moreover, EVDI++ assumes precise spatial-temporal alignment between event streams and frames. While hybrid multi-camera systems can achieve satisfactory alignment through calibration, residual misalignment in scenes with complex depth variations adversely affects performance (\cref{limitation} (c)). Future research will focus on incorporating complementary RGB information and developing alignment-adaptive architectures to address these limitations.


\section{Conclusion}
\re{In this paper, we propose a novel EVDI++ framework that unifies motion deblurring, frame interpolation, and blurry video enhancement within a self-supervised learning paradigm. Specifically, a learnable double integral network is designed to estimate the temporal mapping from reference frames to latent sharp images. A learning-based division module is introduced to recover both sharp and re-blurred images across varying exposure intervals. Furthermore, an adaptive parameter-free fusion module is proposed to generate intermediate results by fusing $\bar{L}_{i}(m)$ and $\bar{L}_{i+1}(m)$ in a learnable-free and temporally adaptive manner. To enable unsupervised learning in real-world scenarios, we develop a set of self-supervised loss functions—namely $\mathcal{L}_{S\text{-}B}$, $\mathcal{L}_{B\text{-}E}$, and $\mathcal{L}_{B\text{-}B}$—which facilitate robust training without requiring labeled data. Compared with its predecessor, EVDI, the proposed EVDI++ achieves superior performance and efficiency across multiple benchmarks. Moreover, EVDI++ offers practical benefits by unifying MD, FI, and BEV tasks into a single framework. This design reduces computational redundancy, simplifies deployment, and supports real-time multi-task inference on resource-constrained platforms such as edge devices and mobile robots, making it well-suited for real-world video restoration applications.}

\ifCLASSOPTIONcaptionsoff
  \newpage
\fi

\bibliographystyle{IEEEtran}
\bibliography{egbib}

\vfill

\vspace{-10mm}

\begin{IEEEbiography}[{\includegraphics[width=1in,height=1.25in,clip,keepaspectratio, trim={20 40 20 40}]{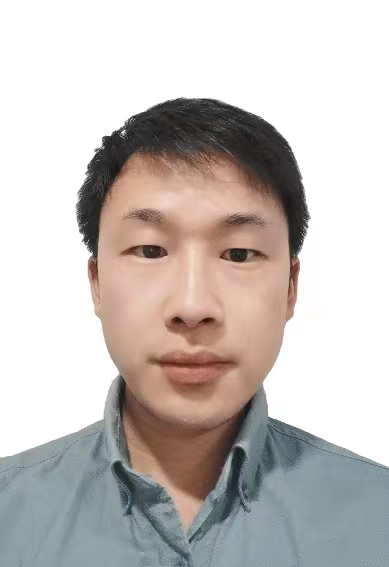}}]{Chi Zhang} is currently a Postdoctoral Researcher at Pengcheng Laboratory, Shenzhen, China. He received the M.S. degree from the School of Information Management at Jiangxi University of Finance and Economics, Nanchang, China, and the Ph.D. degree from the School of Electronic Information at Wuhan University, Wuhan, China. His research interests include image and video processing, as well as neuromorphic vision.
\end{IEEEbiography}

 \vskip -2.5\baselineskip plus -1fil

\begin{IEEEbiography}[{\includegraphics[width=1in,height=1.25in,clip,keepaspectratio, trim={20 40 20 40}]{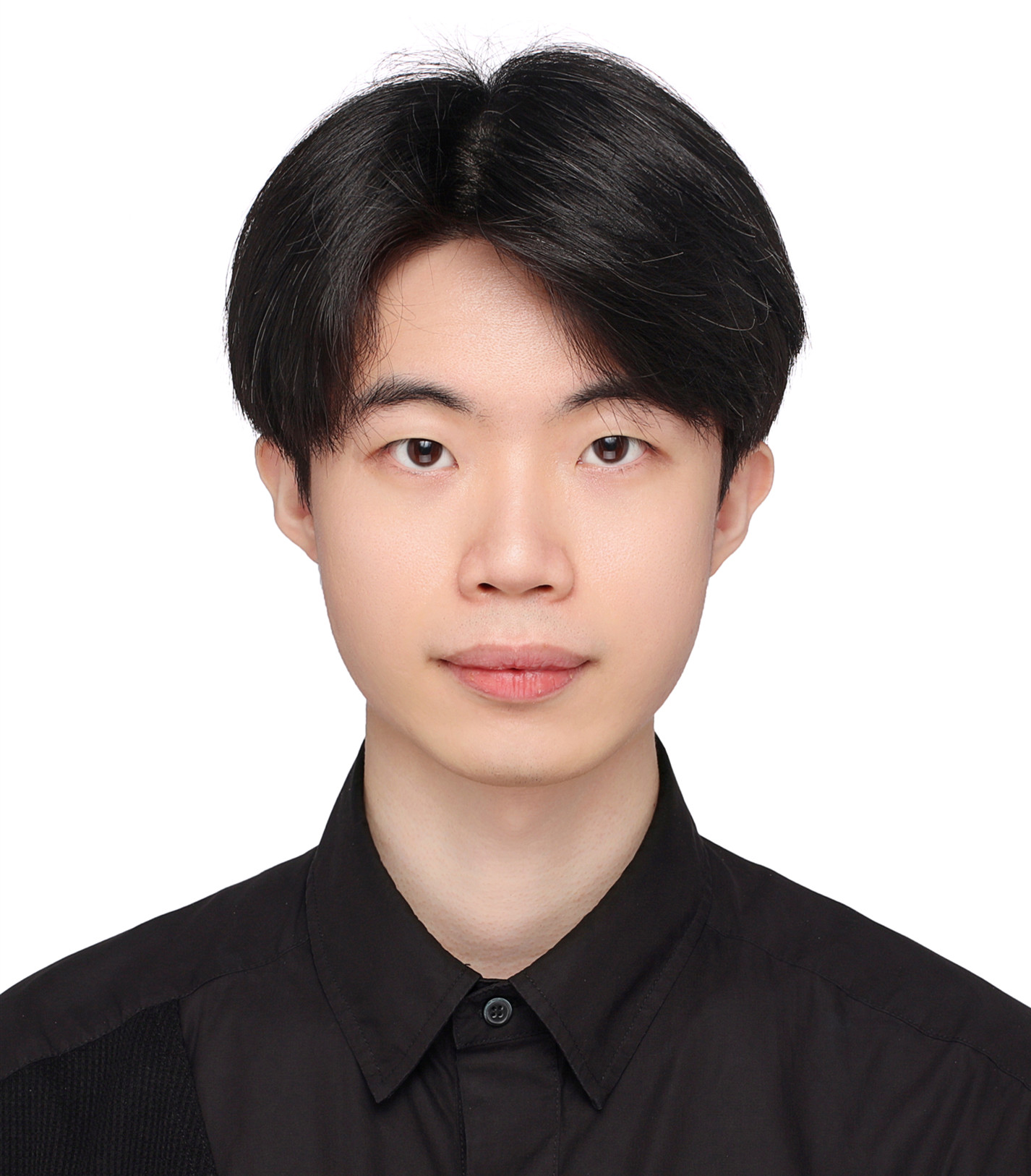}}]{Xiang Zhang}
received his M.E. and B.E. degrees from the Electronic Information School of Wuhan University, Wuhan, China, respectively, in 2023 and 2020. He is currently pursuing a Ph.D. degree in the Department of Information Technology and Electrical Engineering at ETH Zurich, Zurich, Switzerland. His research interests include computer vision and neuromorphic computation.
\end{IEEEbiography}

\vskip -2.5\baselineskip plus -1fil

\begin{IEEEbiography}[{\includegraphics[width=1in,height=1.25in,clip,keepaspectratio, trim={20 40 20 40}]{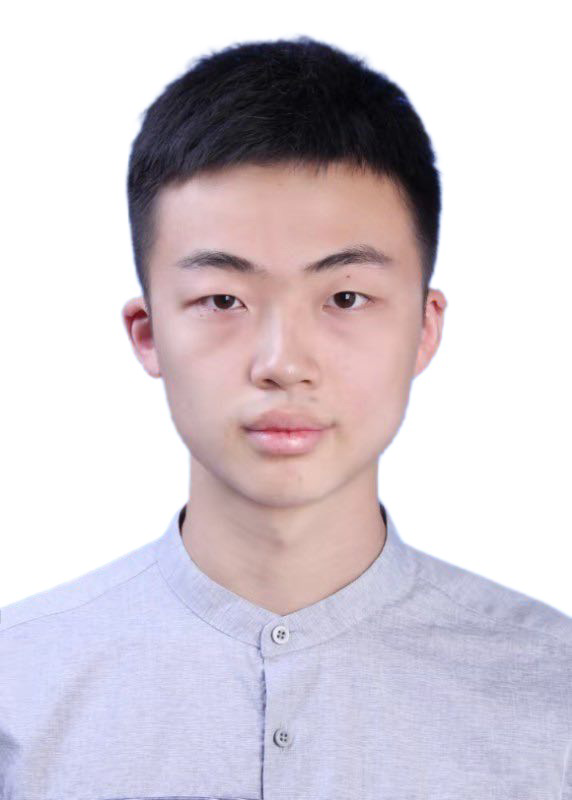}}]{Chenxu Jiang}
received his B.E. degree in electronic information engineering from Wuhan University, Wuhan, China, in 2023. He is currently pursuing an M.E. degree in Information and Communication Engineering at the Electronic Information School, Wuhan University, Wuhan, China. His research interests include computer vision and neuromorphic computation.
\end{IEEEbiography}

\vskip -2.5\baselineskip plus -1fil

\begin{IEEEbiography}[{\includegraphics[width=1in,height=1.25in,clip, keepaspectratio]{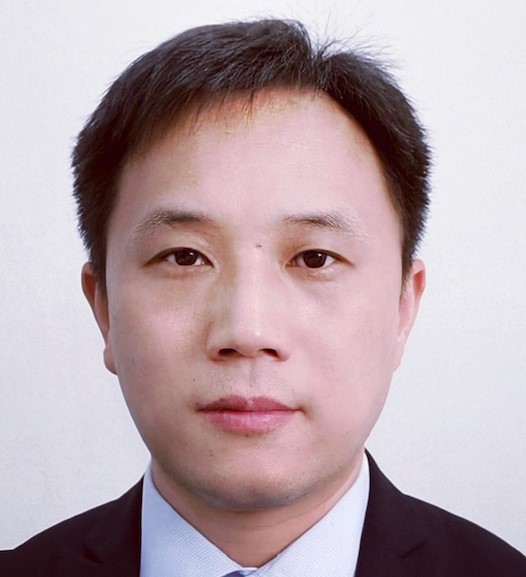}}]{Gui-Song Xia}
is a Professor at the School of Artificial Intelligence and the School of Computer Science, Wuhan University, China. He received his B.S. degree in Electronic Engineering and his M.S. degree in Signal Processing from Wuhan University in 2005 and 2007, respectively, and his Ph.D. degree in Image Processing and Computer Vision from the CNRS Information Processing and Communications Laboratory at Télécom ParisTech, Paris, France, in 2011. His research interests include artificial intelligence, computer vision, photogrammetry, remote sensing, and robotics. He is a Senior Member of IEEE.

\end{IEEEbiography}

\vskip -2.5\baselineskip plus -1fil

\begin{IEEEbiography}[{\includegraphics[width=1in,height=1.25in,clip, keepaspectratio, trim={0 50 0 90}]{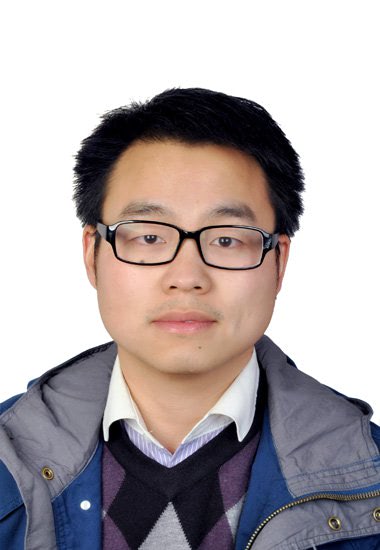}}]{Lei Yu}
 is a Full Professor at the School of Artificial Intelligence, Wuhan University, China. He received his B.S. and Ph.D. degrees in Signal Processing from Wuhan University in 2006 and 2012, respectively. From 2013 to 2014, he was a Postdoctoral Researcher with the VisAGeS Group at the Institut National de Recherche en Informatique et en Automatique (INRIA), France. He has also held visiting positions at Duke University, USA (2016–2017), and at the École Nationale Supérieure de l'Électronique et de ses Applications (ENSEA), Cergy, France (2018). His research interests include artificial intelligence, computer vision, and neuromorphic vision.

\end{IEEEbiography}


\end{document}